\documentclass[lettersize,journal]{IEEEtran}
\usepackage{booktabs}
\usepackage{multirow}
\usepackage{algorithm,algorithmic}

\usepackage{float}
\usepackage{cite}
\usepackage{graphicx}
\usepackage{amsmath,amsfonts}
\usepackage{verbatim}
\usepackage{stfloats}
\usepackage{textcomp}
\usepackage{array}
\usepackage{url}
\usepackage[caption=false,font=normalsize,labelfont=sf,textfont=sf]{subfig}

\begin{document}

\title{Membership Inference Attack with Partial Features}

\author{
Xurun Wang,
Guangrui Liu,
Xinjie Li,
Haoyu He,
Lin Yao,
Zhongyun Hua,
and Weizhe Zhang

\thanks{
Xurun Wang, Guangrui Liu, Xinjie Li are with the School of Cyberspace Science, Harbin Institute of Technology, Harbin, Heilongjiang 150001, China (e-mail: wangxr@stu.hit.edu.cn; grliu@hit.edu.cn; lixinjie@stu.hit.edu.cn).
}
\thanks{Haoyu He is with Department of Data science and Al, Faculty of IT, Monash University, Melbourne, Australia (e-mail: haoyu.he@monash.edu).}
\thanks{Lin Yao is with the School of Software, Dalian University of Technology,
Dalian 116022, China (e-mail: yaolin@dlut.edu.cn).}
\thanks{Zhongyun Hua is with the School of Computer Science and Technology, Harbin Institute of Technology, Shenzhen, Guangdong 518055, China (e-mail: huazhongyun@hit.edu.cn).}
\thanks{Weizhe Zhang is with the School of Cyberspace Science, Harbin Institute of Technology, Harbin, Heilongjiang 150001, China, and also with the School
of Cyberspace Science, Harbin Institute of Technology, Shenzhen, Guangdong 518055, China (e-mail: wzzhang@hit.edu.cn).}
}

\maketitle
\newtheorem{define}{Definition}

\begin{abstract}
Machine learning models are vulnerable to membership inference attack, which can be used to determine whether a given sample appears in the training data. Most existing methods assume the attacker has full access to the features of the target sample. This assumption, however, does not hold in many real-world scenarios where only partial features are available, thereby limiting the applicability of these methods. In this work, we introduce Partial Feature Membership Inference (PFMI), a scenario where the adversary observes only partial features of each sample and aims to infer whether this observed subset was present in the training set.
To address this problem, we propose MRAD (Memory-guided Reconstruction and Anomaly Detection), a two-stage attack framework that works in both white-box and black-box settings. In the first stage, MRAD leverages the latent memory of the target model to reconstruct the unknown features of the sample. We observe that when the known features are absent from the training set, the reconstructed sample deviates significantly from the true data distribution. Consequently, in the second stage, we use anomaly detection algorithms to measure the deviation between the reconstructed sample and the training data distribution, thereby determining whether the known features belong to a member of the training set. Empirical results demonstrate that MRAD is effective across various datasets, and maintains compatibility with off-the-shelf anomaly detection techniques. For example, on STL-10, our attack exceeds an AUC of around 0.75 even with 60\% of the missing features. 
\end{abstract}


\begin{IEEEkeywords}
Membership inference, machine learning, data privacy
\end{IEEEkeywords}

\section{Introduction} \label{sec:intro}
\IEEEPARstart{M}{achine} learning has been widely applied across various domains, such as medical diagnosis\cite{bilionis2025disparate}, traffic analysis\cite{hui2021trajnet}, and autonomous driving\cite{chen2024end}. However, current machine learning models remain vulnerable to privacy inference attacks. Adversaries can exploit interactions with the target model to infer privacy information from its training data, posing significant threats to user confidentiality and data privacy. A well-known example of such attacks is the Membership Inference Attack (MIA)\cite{shokri2017membership, carlini2022membership}, which can be used to determine whether certain samples were included in the training set of a target model. This type of attack highlights the risk of information leakage during inference, making the deployment of machine-learning services more challenging.
\IEEEpubidadjcol

Membership inference attack has been shown to compromise the privacy of various machine learning models\cite{fu2025unlocking, hu2025unveiling ,galichin2025glira,chi2024shadow, wang2023link,zarifzadeh2024low}. As the majority of existing studies focus on classification models, our work follows the same line of research. Most membership inference attacks infer membership by computing attack signals, such as the sample loss, based on feedback from the target model together with prior knowledge of the model structure and training data\cite{song2021systematic,ye2022enhanced, liu2023membership}. Under weaker attacker assumptions, some attacks are designed for settings with limited observability, where the attacker cannot obtain full model outputs\cite{peng2024oslo,li2021membership} or lacks knowledge of the training data distribution and the target model details\cite{liu2024gradient}. These settings further expand the practical scope of membership inference attacks. However, all of these methods share the same goal: determining whether a given sample exactly matches one of the training samples. Recently, Tao et al.\cite{tao2025range} introduced range-based MIA as a new type of membership inference attack that breaks away from this exact-match paradigm. Instead of identifying whether a specific sample appears in the training set, their method aims to infer whether the training data contains any sample whose features fall within a specified range, highlighting that membership inference can extend beyond traditional exact-match assumptions.

\begin{figure}[!ht]
    \centering
    \includegraphics[width=0.7\linewidth]{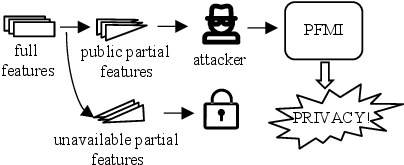}
    \caption{Our goal is to infer membership information when only partial features of a sample are available.}
    \label{fig:scenario}
\end{figure}

Inspired by Tao et al.\cite{tao2025range}, we pose a question: \textit{How much private information can membership inference attack extract when some features are missing?} 
This setting aligns more closely with practical scenarios, as privacy constraints or instability in data sources may prevent an attacker from obtaining complete feature information for the target sample, leaving only partially observed tabular or image data available. To illustrate this, consider an attacker attempting to determine whether an individual, Alice, has a particular disease by querying a diagnostic system deployed in a hospital she frequently visits. If Alice is indeed diagnosed with the disease, it is likely that her data is included in the training set. However, due to the difficulty of acquiring complete information, the attacker may only know a subset of her attributes, such as age, gender, and race. Applying a standard membership inference attack in this setting would require exhaustively enumerating all possible combinations of the unknown features, which is computationally infeasible and inefficient. 

As Figure \ref{fig:scenario} shows, in this paper, we consider a setting where the attacker has access to only a subset of a sample’s features and aims to infer whether this partial feature combination was present in the training set of a target model.  We refer to this type of attack as \textit{\textbf{P}artial \textbf{F}eature \textbf{M}embership \textbf{I}nference} (PFMI). In this setting, the observed partial features is termed known features, while the remaining unobserved ones are referred to as unknown features.
The property existence attack, proposed as a generalization of membership inference \cite{chaudhari2023snap}, shares certain similarities with PFMI in that both seek to infer the presence of specific feature combinations. However, in property existence attack, the adversary can directly access complete samples containing the target feature combination \cite{chaudhari2023snap,wang2024property}, whereas PFMI operates under stricter constraints by relying solely on partial observations. Although the attacker’s limited access to only partial feature information makes the inference task more challenging, this constraint also reflects a more realistic threat model. An adversary equipped with PFMI capabilities can still launch effective privacy inference attacks even with incomplete information. This significantly amplifies the risks faced by existing privacy protection mechanisms, which often assume that attackers have access to complete samples.

However, designing such attacks faces two key challenges:
\begin{itemize}
    \item \textbf{Challenge 1}: Traditional membership inference attacks rely on model feedback\cite{carlini2022membership,zarifzadeh2024low}, but incomplete samples cannot be directly input into the model to obtain meaningful responses.
    \item \textbf{Challenge 2}: A new membership evaluation criterion is required to determine feature-level membership, as existing methods are no longer applicable in this setting.
\end{itemize}

To tackle these problems, we propose \textbf{MRAD}—a two-stage attack framework that combines Memory-guided Reconstruction with Anomaly Detection. Specifically, to address the first challenge, the first stage uses the known features as anchors to recover the unknown features based on the representations learned by the model. Based on our observations, reconstructed samples with non-member features often deviate significantly from the original distribution, resulting in anomalous data. We leverage anomaly detection techniques to identify such deviations, thus determining whether the observed partial feature combination exists in the training set, thereby addressing the second challenge. Further details will be provided in Section \ref{sec:intuition}. Our main contributions can be summarized as follows.

\begin{itemize}
    \item  We investigate a more realistic setting for membership inference attack, in which the adversary has access only to partial features of the target sample. We refer to this scenario as PFMI, and we provide a formal definition of the attack framework under this constraint. 
    \item We propose a two-stage attack framework designed for this new inference scenario, which is applicable to both white-box and black-box settings. The framework is modular and can be integrated with a variety of off-the-shelf anomaly detection techniques.
    \item We extensively evaluate our attack framework on both image and tabular datasets, demonstrating its compatibility with various anomaly detection methods. For the STL-10 dataset, our method still exceeds an AUC of 0.75 even when 60\% of the features are missing.
    \item We show how hyperparameter settings and the importance of observed features affect attack performance by experiments. We also present a case study to demonstrate how our attack can lead to privacy leakage in a real world scenario.
\end{itemize}

The remainder of this paper is organized as follows. Section \ref{sec:related_work} reviews related work on membership inference and anomaly detection. Section \ref{sec:problem_setup} introduces the PFMI setting, including the attack objective, adversary capability and attack definition. Section \ref{sec:MRAD} presents the technical details of the proposed two-stage attack framework. Section \ref{sec:sec4} presents experimental results, including studies that validate design intuition, followed by a comprehensive evaluation of the attack performance in various settings. Section \ref{sec:discussion} analyzes the relationship between PFMI and other inference attacks, and outlines directions for future work. Section \ref{sec:conclusion} concludes the paper.

\section{Related Work}\label{sec:related_work}

In this section, we review prior work on membership inference attack. Then we will give a brief introduction on anomaly detection.

\subsection{Membership Inference Attack}
Membership inference attack against machine learning models was first introduced by Shokri et al.\cite{shokri2017membership}. A binary classifier was trained as the attack model using prediction vectors produced by shadow models. To launch the attack, the target model’s prediction on a given input is provided to the attack model, which decides whether the input was included in the training data.  Most follow-up studies adopt a similar setting, where the adversary has access to rich information, such as the full model outputs and the training data distribution. These methods typically rely on the observation that member samples tend to yield lower loss values than non-members \cite{song2021systematic,sablayrolles2019white,carlini2022membership}. Song et al.\cite{song2021systematic} predict membership with a modified loss function. Ye et al.\cite{ye2022enhanced} proposed a membership inference attack that achieves an arbitrary false positive rate by using a non-member set to calibrate the decision threshold. Carlini et al. \cite{carlini2022membership} trained multiple shadow models to examine the loss distributions of samples when they are in and out of the training set, further improving the attack performance under low false positive rates. Instead of using loss values for membership inference, Zarifzadeh et al.\cite{zarifzadeh2024low} computed a likelihood ratio based on the output confidences and used it as the inference metric. 

Another line of work considers membership inference under restricted attacker capabilities. In the label-only setting, the adversary can only observe the predicted label rather than the full confidence vector \cite{li2021membership,choquette2021label,peng2024oslo}. Li et al. \cite{li2021membership} leveraged the fact that member samples are often farther from the decision boundary, while Peng et al. \cite{peng2024oslo} crafted minimally perturbed adversarial examples using a shadow model and inferred membership based on whether the perturbed samples were misclassified by the target model. As the method proposed by Peng et al.\cite{peng2024oslo} relies solely on the final prediction of the target model, it requires only a single query. Another setting is blind membership inference, in which the adversary has access to the full prediction vector of the target sample but lacks knowledge of the true label. Hui et al. \cite{hui2021practical} proposed a blind black-box attack that performs differential comparisons between samples to infer membership without ground-truth guidance. In addition, Liu et al.\cite{liu2024gradient} demonstrate that membership inference remains feasible even when the adversary lacks prior knowledge of both the target model and the training data.

Recently, Tao et al.\cite{tao2025range} proposed a novel form of membership inference attack that shifts the focus from individual samples to feature regions. Rather than determining whether a specific sample was included in the training set, their method aims to infer whether any data point within a given range appears in the training data. To perform the attack, they first sample instances from the specified range, apply standard membership inference techniques to assign scores to these samples, and then aggregate the scores to reach a final inference decision.

\subsection{Anomaly Detection}

Anomalous data refers to samples that deviate from the expected or learned data distribution. Two common categories of such anomalies are concept drift and out-of-distribution (OOD) samples. Concept drift occurs when the test-time data distribution shifts from the training distribution, leading to degraded model performance \cite{wares2019data}. Yang et al. \cite{yang2021cade} detect such drift by modeling input data with autoencoders and computing Median Absolute Deviation(MAD) values between the current and original distributions. Wan et al. \cite{wan2024online} proposed MCD-DD, which uses a dynamically updated encoder and Mahalanobis distance-based criteria to detect distributional changes in time series. OOD detection has also received significant attention, as the presence of OOD samples can bias model behavior. Park et al. \cite{park2023nearest} improved OOD detection by reducing overconfidence in OOD regions. Reiss et al. \cite{reiss2023mean} designed a novel contrastive loss function that enhances contrastive learning for OOD detection. Liu et al. \cite{liu2025detecting} exploited the clustering effect in the penultimate layer of neural networks to develop a post-hoc OOD detection method.

\section{Problem Formulation}\label{sec:problem_setup}
In this section, we present the problem formulation of the proposed Partial Feature Membership Inference (PFMI) attack. We begin by describing the attack model, followed by a formal definition of the attack.
\subsection{Attack Model}
\textbf{Attack Objective}: In this work, the goal of the attacker is similar to that of traditional MIA, but with a key constraint: the attacker only has access to partial features of the target sample. In this paper, for the sake of generality, we define partial features as a subset of the full feature vector, where the known features can appear at arbitrary positions. That is, the known and unkonwn features are not required to form a contiguous block or follow any fixed pattern. This flexible definition better captures real-world scenarios where feature availability is influenced by privacy constraints or unstable data sources. The objective is to determine whether this particular feature combination exists in the training set of the target model. We refer to such feature combinations as member features if they appear in the training set, and non-member features otherwise. Apparently, any combination drawn from the training set is considered as member features.

\textbf{Adversary Capability}: This work considers both white-box and black-box attack settings. In the white-box setting, the attacker has full access to the target model, including its outputs, training algorithm, internal parameters, and gradients. In contrast, under the black-box setting, the attacker can only obtain the full prediction vector output by the target model. In addition, although the specific target sample may have missing features due to privacy concerns or incomplete collection, we assume that the attacker has access to fully observed auxiliary data drawn from the same underlying distribution as the target model training data. These auxiliary samples are used to support the attack, for example by training proxy models or estimating relevant statistics, which is a common setting in prior membership inference attack settings\cite{shokri2017membership,song2021systematic,ye2022enhanced,zarifzadeh2024low}. This is feasible in practice, as attackers may obtain partial data through social engineering or cyberattacks. However, the attacker is unable to obtain any complete sample that contains the exact feature combination to be inferred.

\subsection{Attack Definition}
Before introducing our proposed attack definition, we first revisit the standard definition of membership inference attack, as formalized in prior work\cite{yeom2018privacy,zarifzadeh2024low} using an indistinguishability game.

\begin{define} 
\label{def:MIA}
(Membership Inference Attack). Let $\pi$ represent the underlying data distribution, $\theta$ the parameters of the target model, $\mathcal{T}$ the training algorithm, and $\mathcal{A}$ the attack algorithm. 
\begin{enumerate}
    \item[\uppercase\expandafter{\romannumeral1}] The Challenger samples a dataset $\mathcal D \sim\pi$ to train a model $\theta\leftarrow\mathcal T (\mathcal D)$
    \item[\uppercase\expandafter{\romannumeral2}] The Challenger flips a fair coin $b$, if $b=1$, then $x\sim \mathcal D$, otherwise $x\sim\pi/\mathcal D$. The Challenger sends the model $\theta$ and the target sample $x$ to the Adversary
    \item[\uppercase\expandafter{\romannumeral3}] The Adversary gets the prediction $\hat{b}\gets \mathcal{A} (\theta,x\odot M,\pi)$
    \item[\uppercase\expandafter{\romannumeral4}] If $\hat{b}=b$, output 1. Otherwise, output 0.
\end{enumerate}
\end{define}

Previous membership inference attack generally assume that the adversary has access to complete feature information of a sample. In contrast, we consider a more realistic and constrained setting where only partial feature values are available. To formalize this, we introduce a binary mask $M$ that indicates which features are known and which are unknown. Using this mask-based representation, we extend the standard indistinguishability framework and define the Partial Feature Membership Inference (PFMI) problem.
\begin{define} 
\label{def:PFMI}
(Partial Feature Membership Inference). Let $\pi$ represent the underlying data distribution, $\theta$ the parameters of the target model, $\mathcal{T}$ the training algorithm, and $\mathcal{A}$ the attack algorithm. We introduce a feature mask $M \in \{0, 1\}^d$, where $M_i = 1$ indicates that the $i$-th feature is known to the attacker, and $M_i = 0$ indicates it is unknown. We use $\odot$ to denote element-wise product.
\begin{enumerate}
    \item[\uppercase\expandafter{\romannumeral1}] The Challenger samples a dataset $\mathcal D \sim\pi$ to train a model $\theta\leftarrow\mathcal T (\mathcal D)$
    \item[\uppercase\expandafter{\romannumeral2}] The Challenger flips a fair coin $b$, if $b=1$, then $x\sim \mathcal D$, otherwise $x\sim\pi/\mathcal D$ such that $x\odot M\notin \mathcal D \odot M$.
    \item[\uppercase\expandafter{\romannumeral3}] The Challenger gets the known features $x_{kno}\leftarrow x\odot M $. Let $\pi'\gets \pi/\{x': x'\odot M=x_{kno}\} $, then the Challenger sends $(\theta$,$x_{kno},\pi')$  to the Adversary.
    \item[\uppercase\expandafter{\romannumeral4}] The Adversary gets the prediction $\hat{b}\gets \mathcal{A} (\theta,x\odot M,\pi)$
    \item[\uppercase\expandafter{\romannumeral5}] If $\hat{b}=b$, output 1. Otherwise, output 0.
\end{enumerate}
\end{define}

\section{MRAD: Memory-Guided Feature Reconstruction \& Anomaly Detection}\label{sec:MRAD}
In this section, we first introduce the underlying design intuition of our attack, and then detail the implementation process of the MRAD framework.

\subsection{Design Intuition}\label{sec:intuition}
In designing our attack, we build on a well‑established observation: overfitted models memorize their training data, causing training members to occupy regions of the input space associated with low loss values\cite{shokri2017membership,salem2023sok,hu2022membership}.  Specifically, for a supervised machine learning algorithm $\mathcal{T}$, the training process aims to learn a model that can effectively distinguish between samples from different classes. Since it is impractical to access the full data distribution, a finite subset of labeled examples $\mathcal D_{train}$, the training set, is typically sampled to represent it. A function $f_\theta$, parameterized by $\theta$, is then learned to map each input $x$ from $\mathcal D_{train}$ to its corresponding label $k$ by minimizing a loss function $\mathcal{L}_\theta(x, k)$. Because the optimization only considers the training set, the model tends to assign lower loss values to training samples compared to unseen ones. This results in loss valleys around training points in the input space, a phenomenon that forms the basis of many membership inference attack \cite{yeom2018privacy,hu2022membership}.

However, as discussed in Section \ref{sec:intro}, Designing a membership inference attack under partial feature knowledge introduces several challenges. We next provide a detailed elaboration of these challenges and their corresponding solutions.

\textbf{Challenge 1}:  Incomplete inputs cannot be fed into the neural network directly, which makes it hard to get feedback from the model. The first step is therefore to determine how to fill in the missing features. Several studies have demonstrated that model inversion attacks can successfully recover input features\cite{dibbo2023model,kahla2022label,an2022mirror,annamalai2024linear,mehnaz2022your,kabir2025disparate}.  
These attacks can be broadly categorized into two types.
The first type aims to reconstruct representative features of an entire class, typically in scenarios where all samples from that class correspond to the same entity \cite{dibbo2023model,kahla2022label,an2022mirror}, and these attacks are generally conducted without prior knowledge of partial features. For example, in a facial recognition system, Fredrikson et al. \cite{fredrikson2015model} introduced the Face‑Rec attack, which reconstructs a human face using the target person’s label. The label, representing the target identity in the recognition system, serves as a guiding signal to steer the reconstruction toward the correct person. However, such attacks are not directly applicable to our setting, since in our case a class label does not uniquely correspond to a single entity. The second type of attack, also known as attribute inference attack, attempts to recover unknown private attributes based on the known features of a sample\cite{mehnaz2022your,annamalai2024linear,kabir2025disparate}. Yet, most existing works in this category assume that the attacker has prior knowledge of the possible values of the missing features, which is unrealistic in practice.

\textbf{Solution to Challenge 1}: We adopt a reconstruction strategy similar to that of Fredrikson et al. \cite{fredrikson2015model}, leveraging the model’s inherent memorization of its training data. Specifically, instead of using labels, we treat the known features as anchors and apply backpropagation to optimize the unknown features, aiming to minimize the sample’s loss value. Our goal is to recover what the model “remembers” , that is, the internal impression it has formed during training, typically corresponding to a point near a local minimum of the loss function. Due to model memorization, the differences present in reconstructed samples serve as crucial evidence for inferring membership.

\textbf{Challenge 2}: Membership inference becomes more challenging when feature representations are ambiguous. Multiple samples may share the same known features, instead of reproducing the true underlying sample, the reconstruction often collapses into a blended or averaged representation shaped by several plausible candidates. This prevents the model from faithfully capturing the characteristics of any single instance, causing the reconstructed result to deviate noticeably from the original. As a consequence, membership inference methods that rely on detecting exact matches become fundamentally ineffective.

Moreover, since the attacker never observes a fully specified instance with the target feature combination, there is no principled way to characterize the feasible space of the unknown attributes. Without a well-defined feature range, it is not possible to establish meaningful boundaries for sampling, effectively ruling out the use of range-based attack proposed by Tao et al.\cite{tao2025range} in this setting.

\textbf{Solution to Challenge 2}: We shift our focus from exact sample matching to assessing how well the reconstructed samples align with the underlying data distribution. According to our observations, when the known features come from a member sample, the reconstruction tends to better align with the true data distribution, as the model has already formed internal representations based on the complete training inputs. In contrast, if the known feature combination is not a member, the optimization may converge to a random local minimum, resulting in a reconstruction that deviates more from the training distribution. Based on this observation, we employ anomaly detection techniques to identify reconstructed samples that deviate from the true data distribution, enabling us to distinguish between member and non-member feature combinations. The experiment results in Section \ref{sec:performance_eval}  confirm the validity of our observations.

\begin{figure*}[!ht]
    \centering
    \includegraphics[width=0.8\linewidth]{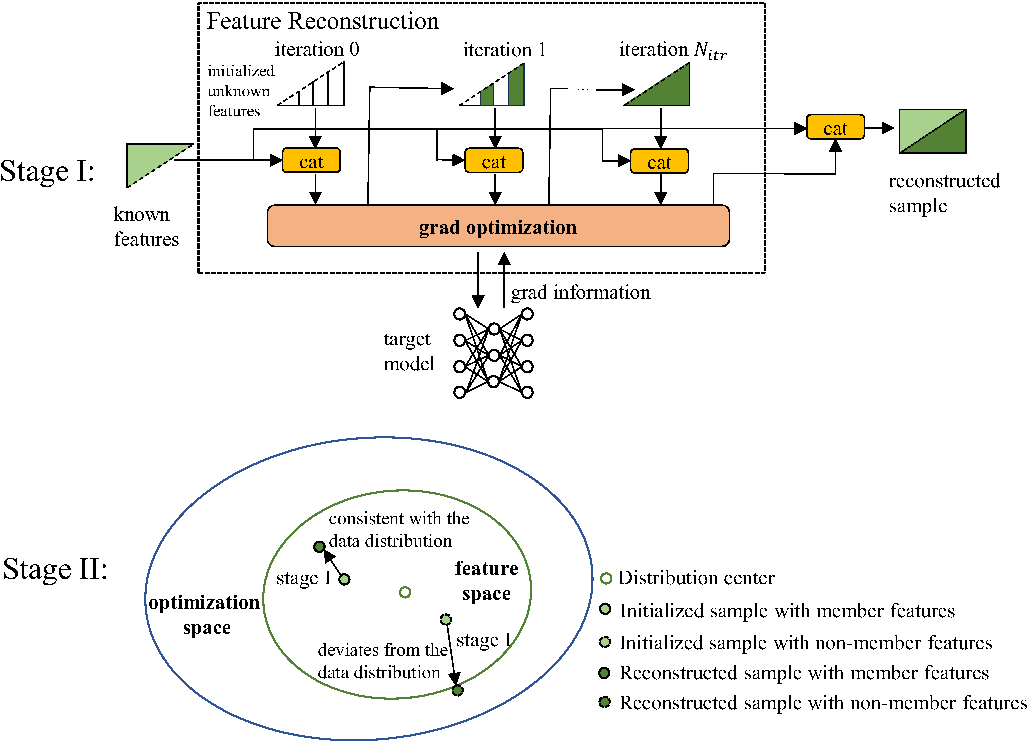}
    \caption{Attack framework:We first implement a simple feature reconstruction algorithm to obtain a complete sample, and then distinguish between member and non-member features based on their deviation distance. }
    \label{fig:framework}
\end{figure*}

\subsection{Detailed Attack Procedure}
As shown in Figure \ref{fig:framework}, to perform a partial membership inference attack, we first leverage the target model to conduct memory-guided reconstruction of the unknown features. This process can be formalized as
\begin{equation}
\min_{x_{unk}}\,\, \mathcal L_\theta(\text{cat}(x_{kno},x_{unk},M),k), 
\end{equation}
where cat denotes the feature concatenation function that merges the known features $x_{kno}$ and the unknown features $x_{unk}$ according to their original feature order. The mask $M$ serves as a positional indicator, specifying which elements in the input vector correspond to known or unknown features. we adopt the loss function $\mathcal L_\theta$ used during model training as our optimization objective. Since multiple samples from different classes may share the same known features, the label $y$ can be arbitrarily chosen from among those associated with the known feature values. We then apply anomaly detection techniques to determine whether the reconstructed sample $\hat{x}$ deviates from the true distribution of class-$k$ samples. Our framework is compatible with any existing anomaly detection method, and we also propose a simple yet effective algorithm. The remainder of this section introduces our approach from two perspectives: memory-guided feature reconstruction and anomaly detection.

\textbf{Memory-Guided Feature Reconstruction}: As shown in Algorithm \ref{alg:MR}, for each sample with missing feature values, we begin by randomly sampling a label $k$ from label set $K$ (line 1) to account for the fact that a single feature combination may be associated with multiple labels. We then initialize all unknown features to zero (line 2). This zero-initialization ensures that, during the early stages of optimization, the model’s predictions remain primarily influenced by the known features rather than arbitrary initialization noise. Next, we merge the known and initialized unknown features into their original ordering to form a complete input vector (line 3). Since the attacker has access to the loss function used during the model’s training, the unknown features can be iteratively optimized through backpropagation (lines 4–8). By repeatedly updating the unknown features to minimize the sample’s training loss, the procedure gradually reconstructs feature values that are consistent with the target model’s learned decision boundary.

\begin{algorithm}
\caption{Memory-guided Reconstruction}
\label{alg:MR}
\begin{algorithmic}[1]
\REQUIRE Known features $x_{kno}$ from class k, possible label index set $K$, target model $\theta$, number of iterations  $N_{itr}$, step length $\eta$, known feature mask $M$, loss function $\mathcal L$ during the training process of the target model.           
\STATE $k \leftarrow \text{RandomSample}(K)$
\STATE $x_{unk}\gets0$
\STATE $\hat{x}\gets \text{cat}(x_{kno},x_{unk},M)$ // Combine the known and unknown features according to their original feature order.
\FOR{$i=1$ to $N_{itr}$}
\STATE $g\gets grad(\mathcal L_\theta(\hat{x},k))$ 
\STATE $(g_{kno}, g_{unk})\gets \text{split}(g,M)$ // Split the gradient vector into gradients of known features and gradients of unknown features.
\STATE $x_{unk}\gets x_{unk}-\eta\cdot g_{unk}$
\STATE $\hat{x}\gets \text{cat}(x_{kno},x_{unk},M)$
\ENDFOR
\RETURN $\hat{x}$
\end{algorithmic}
\end{algorithm}

\textbf{Anomaly Detection}: As shown in Algorithm \ref{alg:AD}, to assess the degree of deviation of a sample, we first model the distribution of real data (lines 1–8). This involves sampling an auxiliary inference set $\mathcal{D}_{aux}$ from the data distribution and computing the centroid $c_k$ and dispersion for each class. For measuring dispersion, we adopt an approach similar to CADE \cite{yang2021cade}, using the Median Absolute Deviation (MAD) as a robust indicator of spread. Then we compute the distance $\hat{d}$ between the reconstructed sample $\hat{x}_k$ and the centroid of class $k$, and derive the corresponding base deviation distance $\delta$ (lines 9–10). To further enhance inference performance, we train a shadow model $\theta_s$ on data drawn from the same distribution and compute a shadow deviation distance $\delta_s$ on shadow model based on the same known features, following the same reconstruction procedure. In summary, both $\delta$ and $\delta_s$ are derived using identical feature values, but from the target and shadow models respectively. The shadow deviation $\delta_s$ serves as a reference for how typical the deviation would be if the known features is not a member of the training set. Then, we define a membership inference rule based on the relative deviation ${\delta_s}/{\delta}$. Let MRAD denote this relative deviation and the inference rule can be formulated as:
\begin{equation}
\text{MRAD}(x_{kno},\theta,\pi') > \tau,
\end{equation}
If this condition is satisfied, we infer that the known feature combination belongs to a member of the training set,  where $\tau$ is a predefined threshold. By varying $\tau$, we can plot the ROC curve to evaluate the effectiveness of this detection strategy.

\textbf{Threshold Selection}: Previous work has shown that MIA is only meaningful when the false positive rate is sufficiently low\cite{carlini2022membership}. Therefore, we adopt a threshold selection strategy similar to that proposed by Ye et al.\cite{ye2022enhanced}. Specifically, the adversary can sample a batch of samples $\mathcal D_{non}$ from the overall non-member data space $\pi'$. Given a predefined acceptable false positive rate $\alpha$, the adversary then selects a threshold that satisfies 
\begin{equation}
\frac{\lvert\{ x_{kno}\in\mathcal D_{non}\odot M: \text{MRAD}(x_{kno},\theta,\pi')>\tau \}\rvert}{\lvert \mathcal{D}_{non} \rvert}=\alpha.
\end{equation}

\begin{algorithm}
\caption{Anomaly Detection}
\label{alg:AD}
\begin{algorithmic}[1]
\REQUIRE sample $\hat{x}_k$ generated from Algorithm 1 with target model, $\hat{x}_{s,k}$ generated from Algorithm 1 with shadow model, label  $k\in\{1,\dots,K\}$, auxiliary data $x_{aux,k}$ drawn from data distribution, number of samples $n_k$ per class in the auxiliary data.
\FOR {k=1 to  K}
\STATE $c_k \gets \frac{1}{n_k}\sum_{i=1}^{n_k}x_{aux,k}^i$
\FOR {$i=1$ to $n_k$}
\STATE $d_{aux,k}^i\gets \lVert x_{aux,k}^i-c_k\rVert_2$
\ENDFOR
\STATE $\tilde{d}_k \gets \mathrm{median}(d_{aux,k}^i)$, $\,i=1,\dots,n_k$
\STATE $\text{MAD}_k\gets \mathrm{median}(\lvert d_{aux,k}^i-\tilde{d}_k\rvert)$
\ENDFOR
\STATE $\hat{d}\gets \lVert \hat{x}_k-c_k\rVert_2$,  $\hat{d}_s\gets \lVert \hat{x}_{s,k}-c_k\rVert_2$
\STATE $\delta\gets\frac{\lvert\hat{d}-\tilde{d}_k\rvert}{\text{MAD}_k}$, $\delta_s\gets\frac{\lvert\hat{d}_{s}- \tilde{d}_k\rvert}{\text{MAD}_k}$
\RETURN ${\delta_s}/{\delta}$
\end{algorithmic}
\end{algorithm}

\subsection{Black-box Attack} \label{sec:black_box}
The white-box results establish the core feasibility of our approach, but this setting rarely reflects practical deployment environments. To bridge this gap, we develop a black-box variant of our attack. The key idea is to approximate the required gradients using zeroth-order gradient estimation\cite{wibisono2012finite}, allowing the attacker to operate without direct access to the model’s internal parameters or intermediate computations. 

Zeroth-order gradient estimation approximates the gradient via finite differences: small perturbations are applied to the input along multiple random directions $\mu_i$, and the corresponding output variations are used to construct the gradient estimate. Let $m$ denote the number of random directions, $\epsilon$ the perturbation magnitude. $\mathcal L$, as the loss function, serves as the objective function for gradient estimation. Then, the estimated gradient at a $p$-dimensional data point $x$ whose loss under the loss function $L$ can be formulated as 

\begin{equation}
    \nabla \mathcal L=\frac{1}{m}\sum_{i=1}^mp\frac{f(x+\epsilon\mu_i)-f(x)}{\epsilon}\mu_i.
\end{equation}

Since zeroth-order gradient estimation requires drawing multiple samples in the neighborhood of the target point $x$, the black-box attack issues $m+1$ times more queries than its white-box counterpart—$m$ for perturbed inputs and one for the original sample. A very large $m$ not only increases query overhead but also makes the query pattern more likely to be interpreted as unusual or suspicious behavior by the target system, whereas an overly small $m$ leads to a poor approximation of the underlying gradient. Consequently, $m$ acts as a key trade-off variable that balances query efficiency and stealthiness against the fidelity of the estimated gradient and, ultimately, overall attack performance. We will test the black-box setting at section \ref{exp:black_box}.

\section{Evaluation}\label{sec:sec4}
In this section, we evaluate the attack performance of the proposed algorithm. We begin with a quantitative assessment under the white-box setting using four widely adopted datasets. These experiments examine the capability of our attack across different known-feature percentage and include a series of in-depth analyses, such as parameter sensitivity experiments, ablation study, and evaluations of feature-importance effects. We then test the black-box variant of our algorithm on the same four datasets to assess its effectiveness under restricted access conditions. Finally, we demonstrate how our attack operates in real-world scenarios to highlight its practical applicability. 

\subsection{Experiment setup}
\textbf{Dataset \& Model}: We evaluate our attack with three widely used image datasets and one tabular dataset. For all image datasets, the target model is Resnet50, while for the tabular dataset, we employ a multilayer perceptron (MLP). We train each model for 150 epochs, with a learning rate of 0.01 for image datasets and 0.1 for tabular datasets. We provide a detailed overview of each dataset as below.
\begin{itemize}
    \item \textbf{CIFAR-10}\cite{krizhevsky2009learning}: CIFAR-10 is a widely used benchmark dataset for image classification. It contains 50,000 training images and 10,000 test images, each being a 32×32 RGB image.
    \item \textbf{Fashion-MNIST}\cite{xiao2017fashion}: Fashion-MNIST is a drop-in replacement for the original MNIST dataset, featuring 60,000 training images and 10,000 test images. Each image is a 28×28 grayscale representation of an article of clothing.
    \item \textbf{STL-10}\cite{coates2011analysis}: STL-10 contains 13,000 labeled images and 100,000 unlabeled images, making it suitable for supervised, unsupervised, and self-supervised learning. In our experiments, we use only the labeled subset. Each image is a 96×96 RGB image.
    \item \textbf{Epsilon}\cite{pascal_large_scale}: Epsilon was derived from the PASCAL 2008 challenge, contains simulated physical data for a binary classification task. Each sample consists of 2,000 normalized numerical features.
\end{itemize}

\textbf{Known Features \& Unknown Features}: We identify unknown features using a randomly generated mask. For image data, the mask is applied over the 2D pixel grid. In the case of a three-channel image, if a pixel is selected as unknown, all three channels at that pixel location are treated as unknown features.

\textbf{Evaluate Metrics}: We evaluate the attack performance using standard metrics commonly used in membership inference research. Specifically, we use the Area Under the ROC Curve (AUC) to measure the overall effectiveness of the attack, and the true positive rate at a false positive rate of 0.1 (TPR@0.1FPR) to assess how well the attack performs under a strict low-false-positive setting.

\textbf{Anomaly Detection Methods}: In addition to our proposed  method, we integrate three representative anomaly detection techniques: CADE \cite{yang2021cade}, MSAD \cite{reiss2023mean}, and NCI \cite{liu2025detecting}. These comparisons demonstrate the compatibility of our framework with various existing techniques and allow for a comprehensive evaluation across different methodologies. The three baseline methods are summarized as follows:
\begin{itemize}
    \item CADE: CADE is a concept drift detection method that leverages contrastive learning with an autoencoder. It minimizes the distance between embeddings of samples from the same class while maximizing the distance between those from different classes, causing distribution-shifted samples to deviate significantly in the embedding space.
    \item MSAD: MSAD is an out-of-distribution (OOD) detection approach. It fine-tunes a pre-trained model using a specially designed contrastive loss and then detects anomalies based on the similarity between a sample and in-distribution samples. In our implementation, we use a locally trained shadow model as the pre-trained backbone for MSAD.
    \item NCI: NCI is a post-hoc OOD detection method based on the phenomenon of neural collapse. It observes that in-distribution (ID) samples tend to align with their class prototype vectors in the penultimate layer, whereas OOD samples do not. Additionally, ID features tend to cluster near the origin. We adopt the publicly available implementation from github\cite{zhang2023openood}.
\end{itemize}

\subsection{Performance Evaluation} \label{sec:performance_eval}
\begin{figure*}
    \centering
    \subfloat[CIFAR-10]{
    \includegraphics[width=0.22\linewidth]{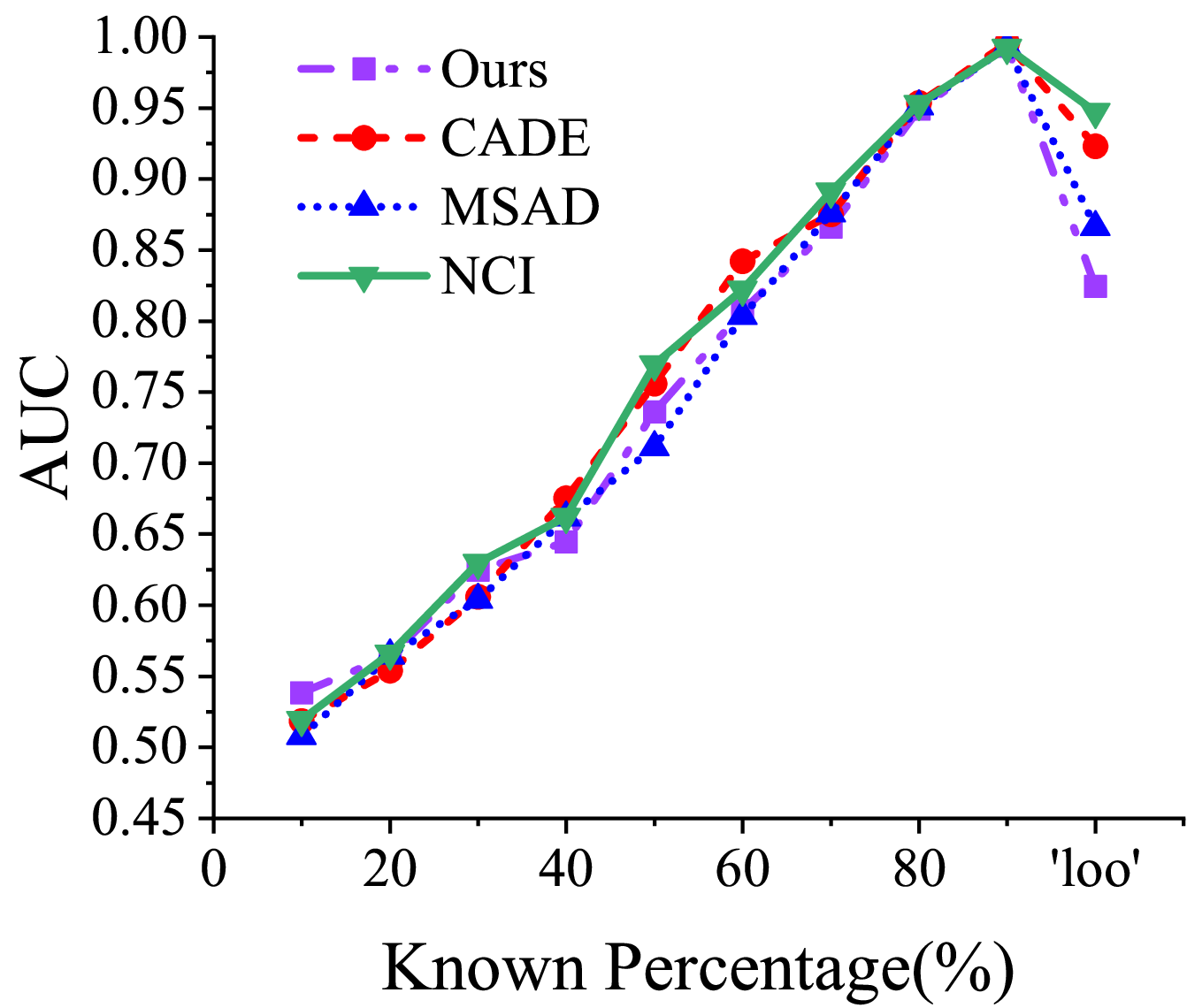}
    }
    \subfloat[STL-10]{
    \includegraphics[width=0.22\linewidth]{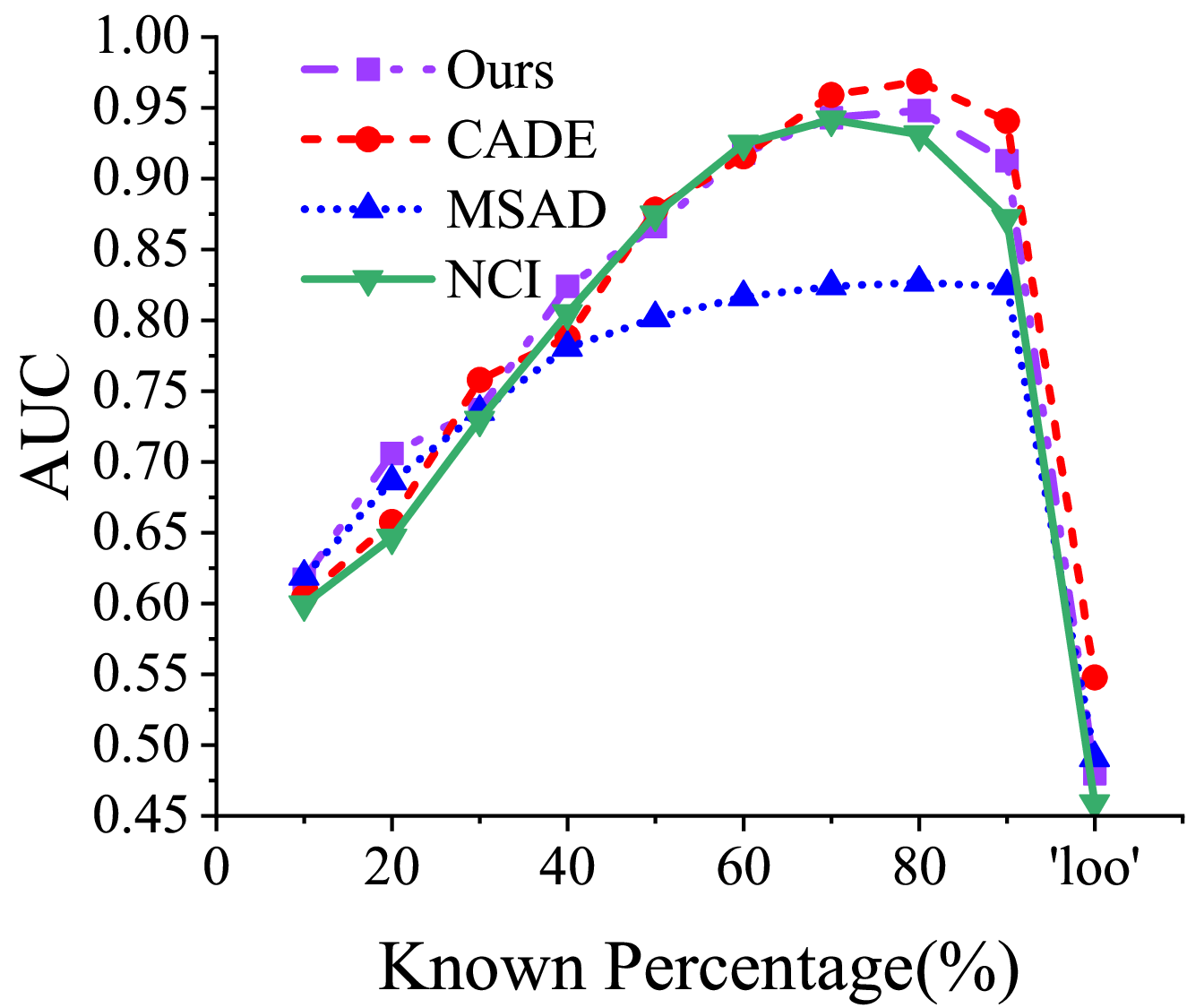}
    }
    \subfloat[Fashion-MNIST]{
    \includegraphics[width=0.22\linewidth]{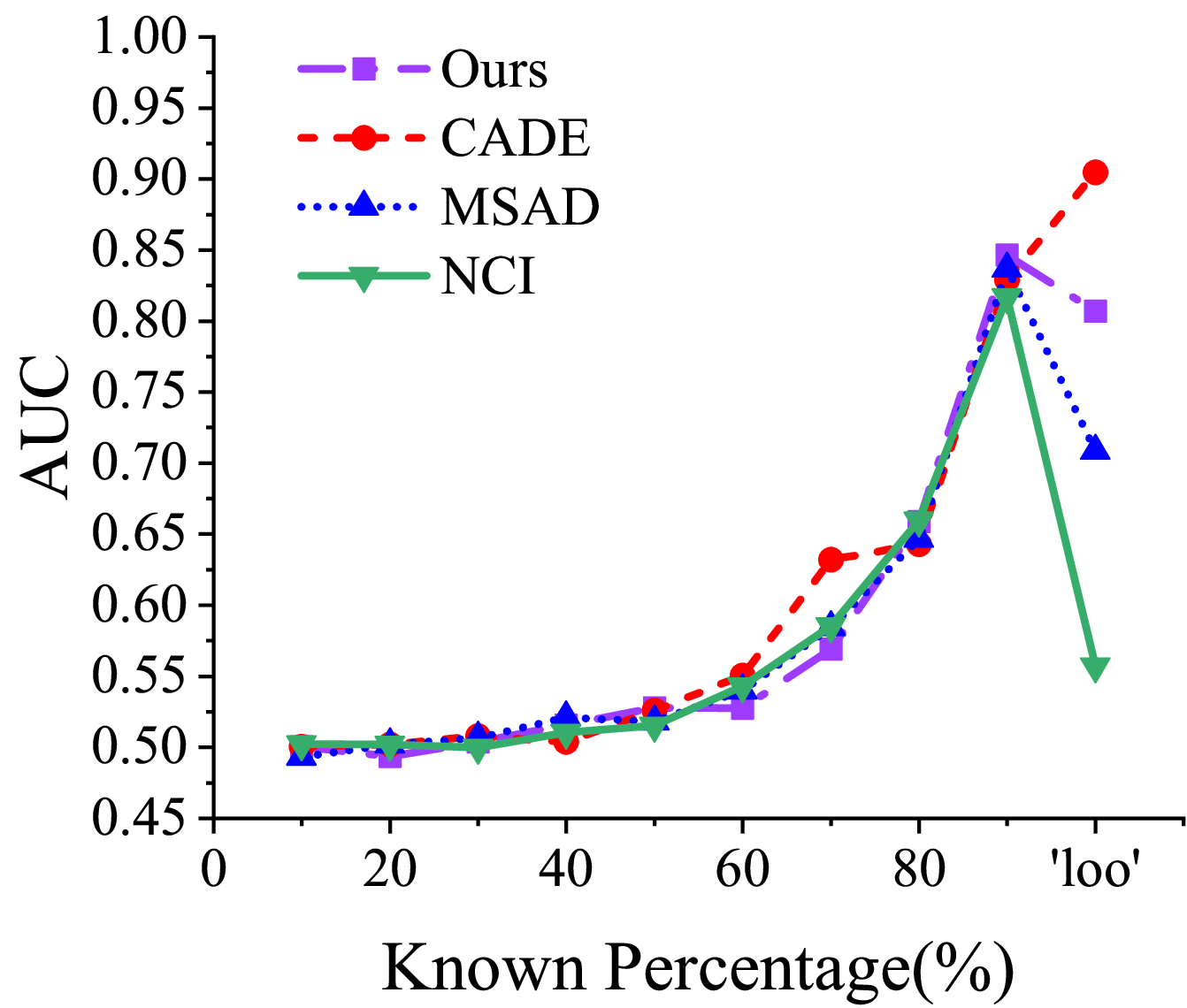}
    }
    \subfloat[Epsilon]{
    \includegraphics[width=0.22\linewidth]{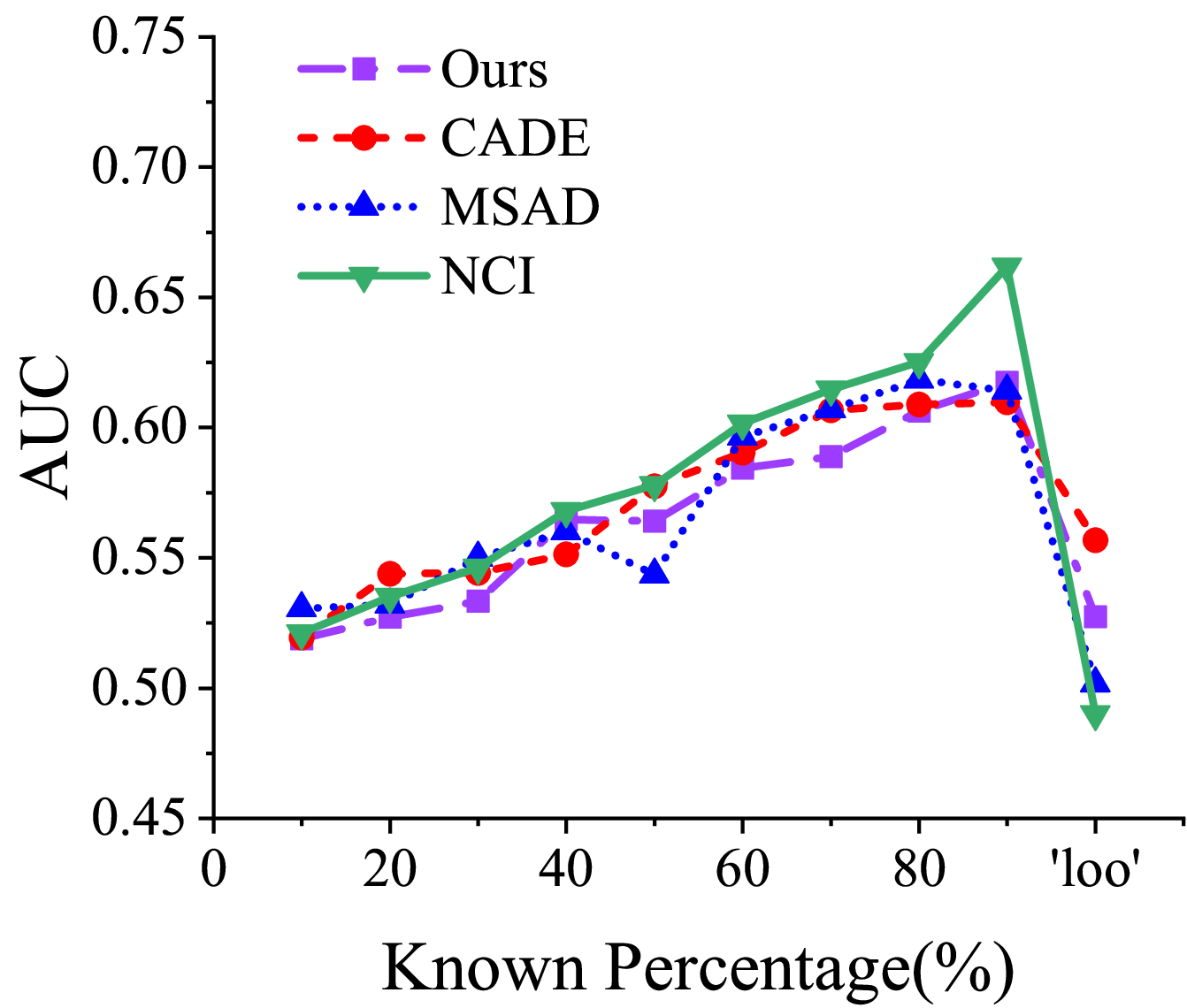}
    }
    \caption{Attack performance (AUC) under varying known feature proportions. The x-axis represents the percentage of known features, and the y-axis shows the corresponding attack AUC. Each curve corresponds to one of the four integrated anomaly detection methods within our attack framework. We denote the case where only one feature is unknown as “loo”.}
    \label{fig:auc_nf}
\end{figure*}

\begin{figure*}
    \centering
    \subfloat[CIFAR-10]{
    \includegraphics[width=0.22\linewidth]{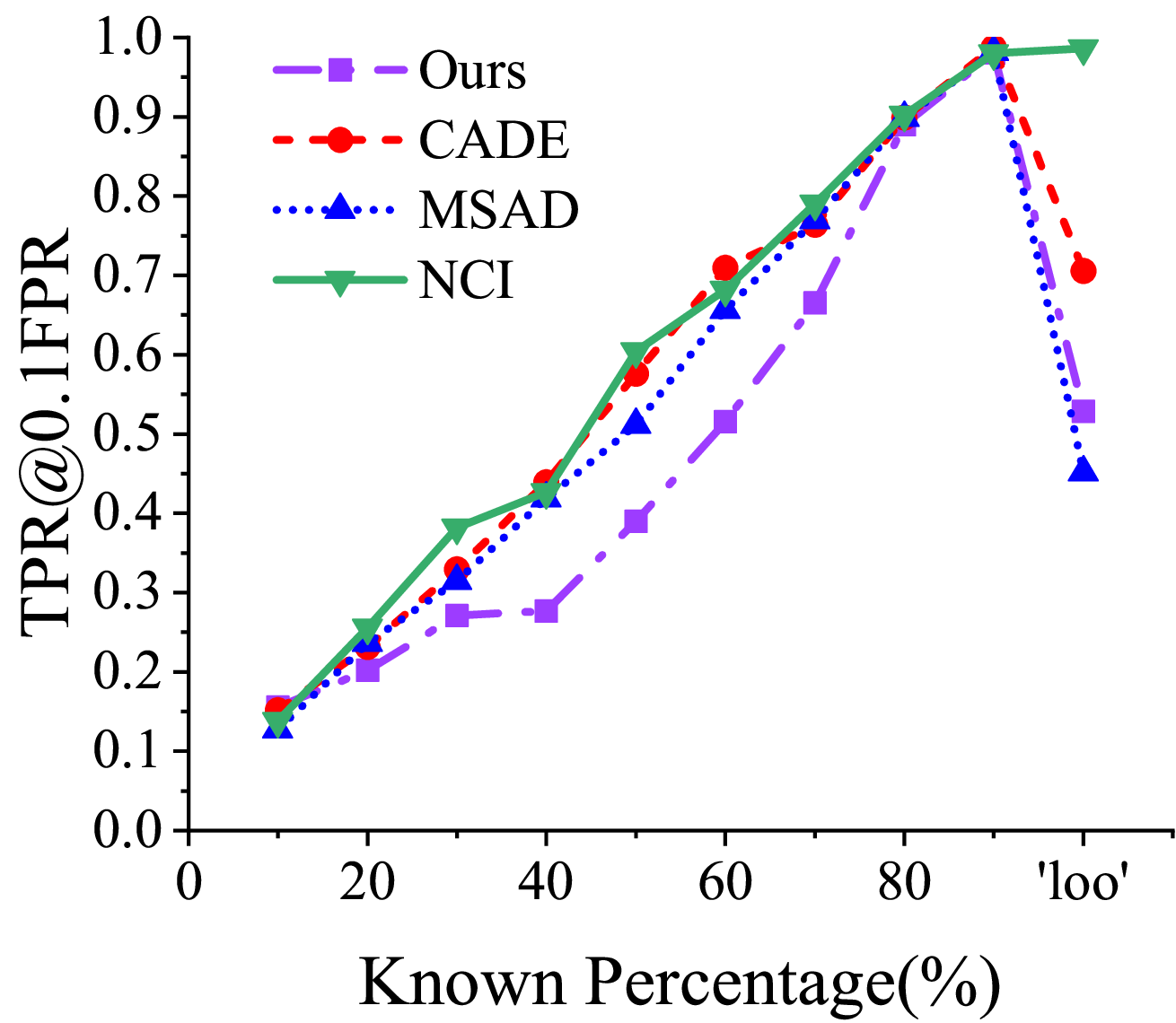}
    }
    \subfloat[STL-10]{
    \includegraphics[width=0.22\linewidth]{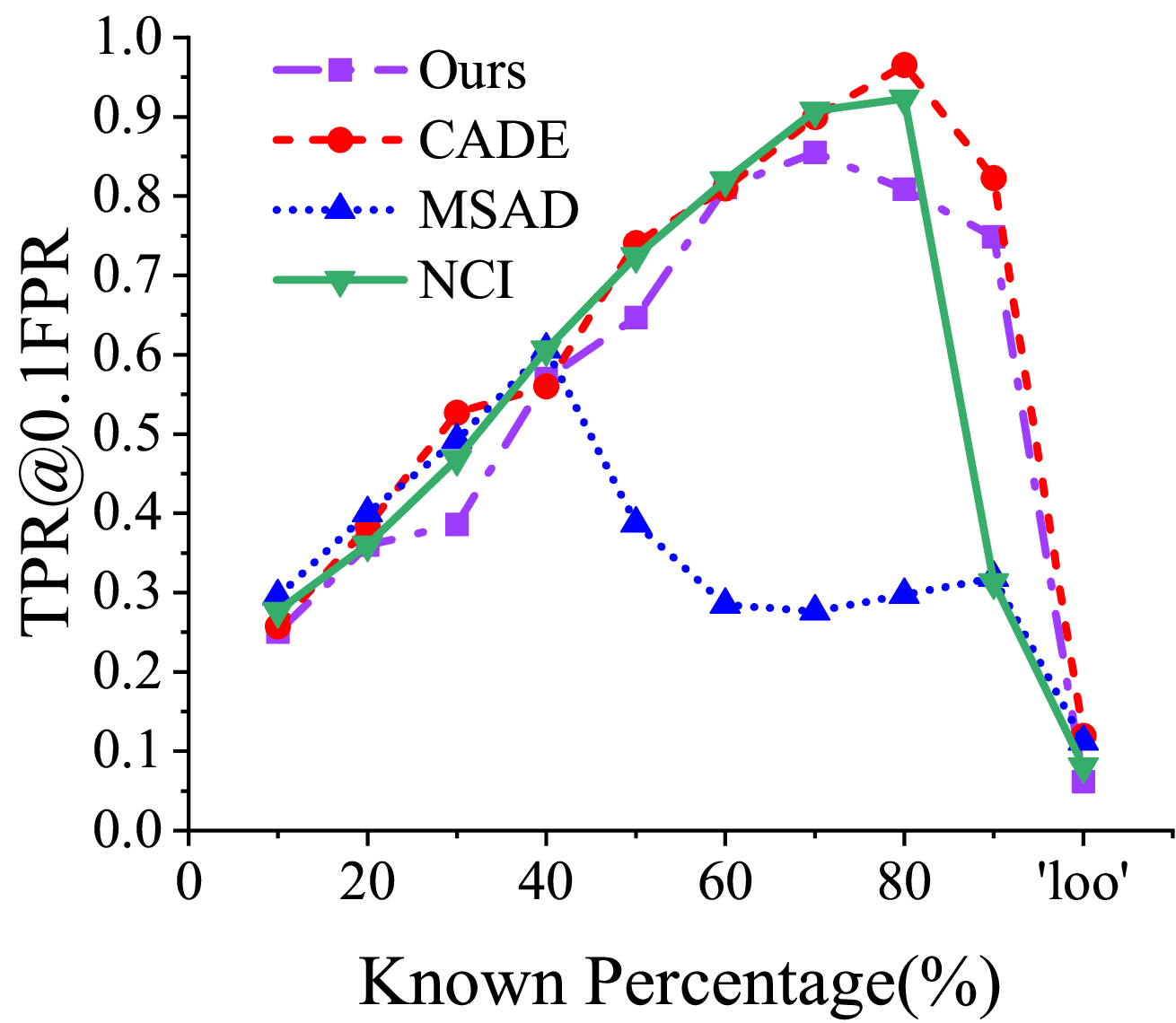}
    }
    \subfloat[Fashion-MNIST]{
    \includegraphics[width=0.22\linewidth]{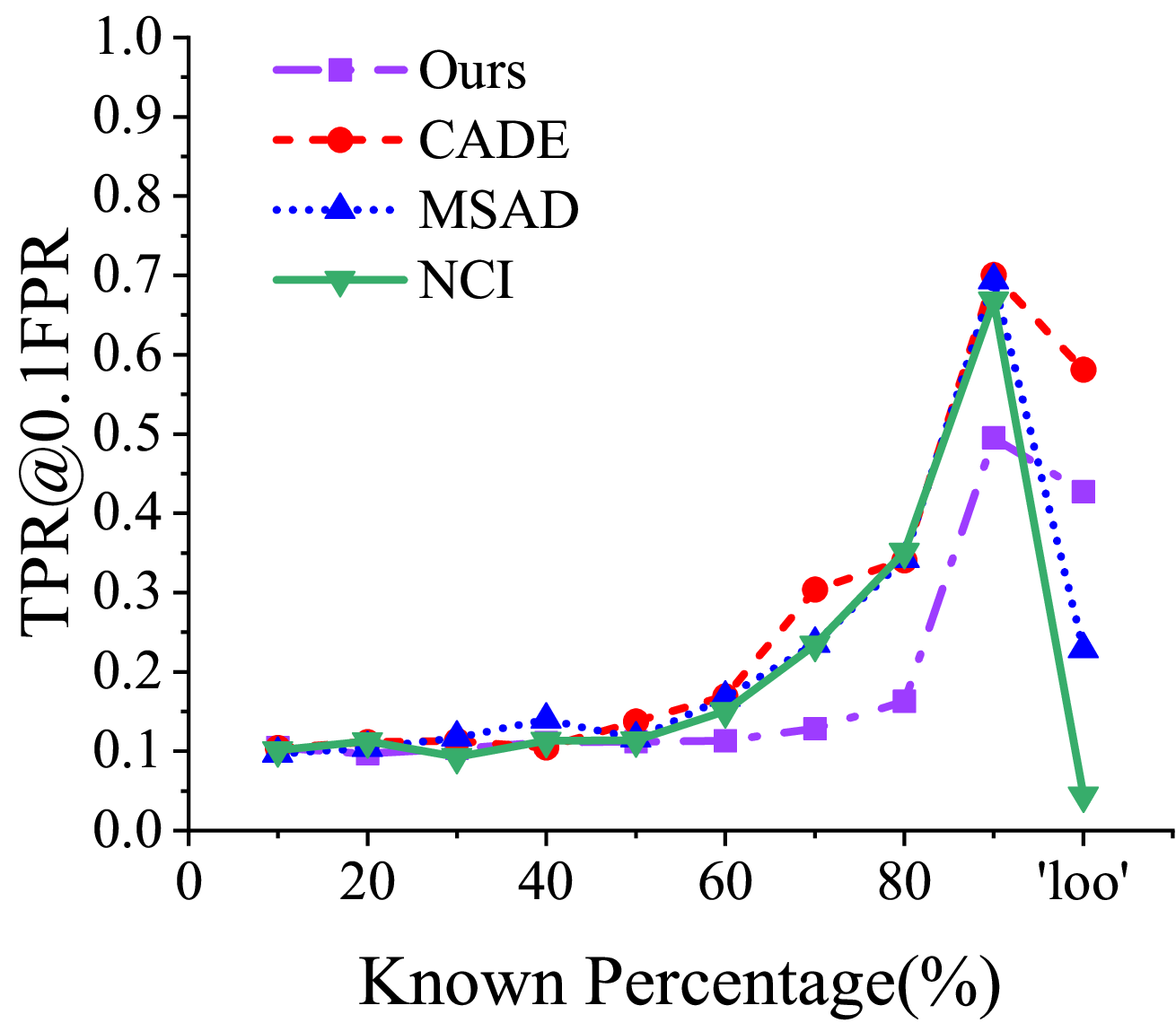}
    }
    \subfloat[Epsilon]{
    \includegraphics[width=0.22\linewidth]{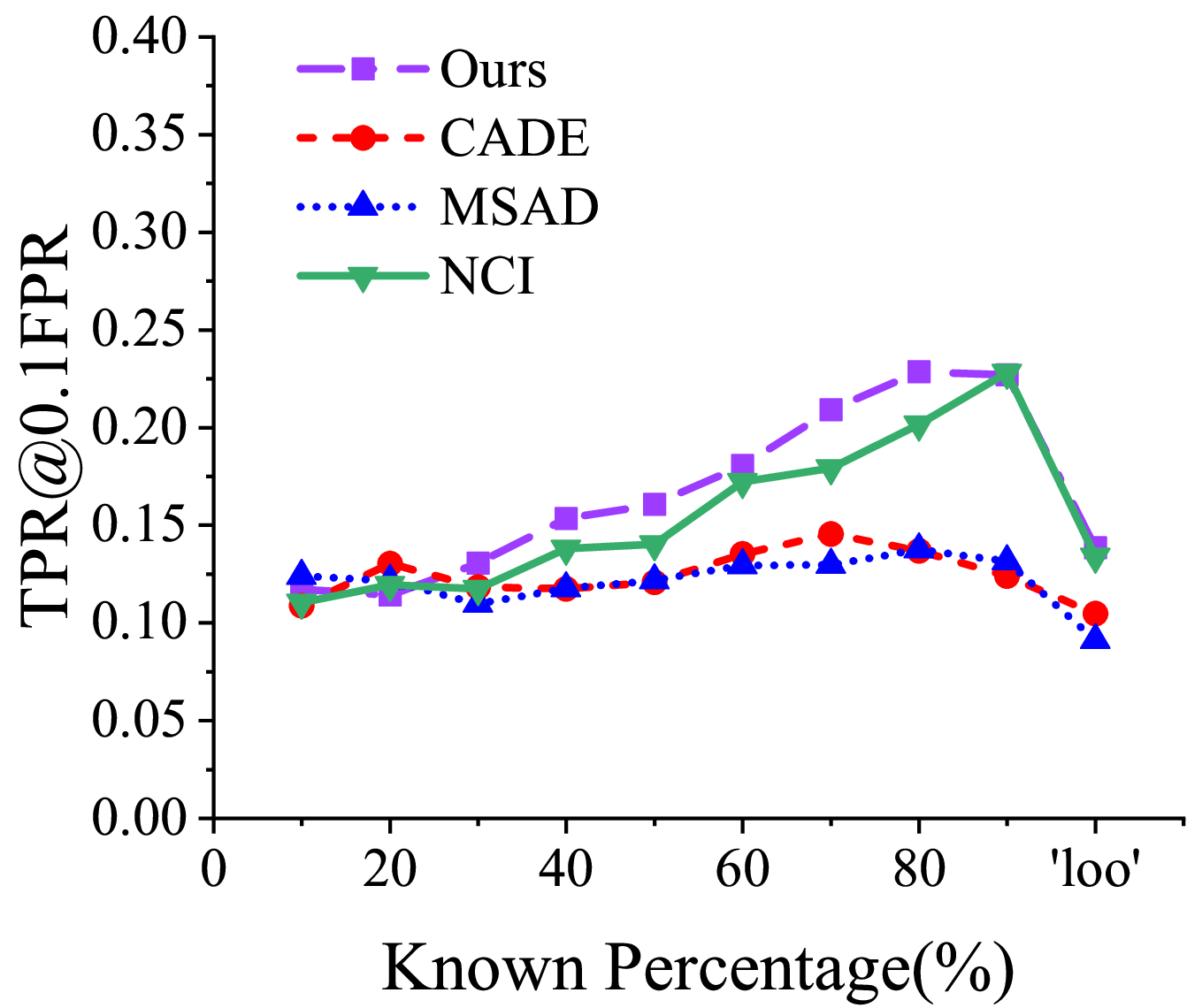}
    }
    \caption{Attack performance (TPR@0.1FPR) under varying known feature proportions. The x-axis represents the percentage of known features, and the y-axis shows the corresponding attack TPR when TPR=0.1. Each curve corresponds to one of the four integrated anomaly detection methods within our attack framework. We denote the case where only one feature is unknown as “loo”.}
    \label{fig:tpr_nf}
\end{figure*}

We begin by evaluating the attack performance under different proportions of known features. We vary the known feature percentage from 10\% to 90\% in increments of 10\% and conduct experiments at each level. In addition, we include a leave-one-out setting where only one feature is missing. This setting is not intended to model a realistic attacker, but rather serves as an extreme and informative limit case of partial-feature knowledge. Because the gradient magnitude is small, we set a larger step size of $\eta = 1000$ and perform only one optimization step (the effect of hyperparameters will be further discussed in Section \ref{sec:sensitivity}). The results are shown in Figures \ref{fig:auc_nf} and \ref{fig:tpr_nf}, where the x-axis indicates the percentage of known features. The leave-one-out setting is marked as "loo" on the x-axis. 

The evaluation results demonstrate a clear pattern. The attack becomes increasingly effective as more features are known to the adversary. On CIFAR-10 and STL-10, the AUC remains above 0.7 even when half of the features are missing, and the attack still achieves a strong TPR. The Epsilon dataset follows a similar trajectory. Although its absolute performance is slightly lower than that of the image datasets, the attack remains consistently effective and well within a practically usable range. When the known-feature ratio reaches 90\%, attack performance peaks on most datasets, with all image datasets achieving an AUC of at least 0.8. This upward trend of attack performance is expected, since additional known features provide tighter constraints for gradient-based reconstruction, enabling the attacker to produce reconstructions that are more consistent with the true data distribution.

However, an important reversal occurs when the known-feature ratio becomes exceedingly high. As more features are revealed, regardless of how poorly the unknown features are reconstructed, the samples increasingly align with the original data distribution. Even when the known features are from a non-member, the reconstructed samples may still evade detection by the anomaly detection algorithm in the second stage. Consequently, the overall attack performance degrades. When the known-feature ratio reaches 100\%, Stage I becomes redundant and the method effectively reduces to a conventional membership inference attack. In that extreme case, standard MIA techniques are typically the more appropriate choice.

\subsection{Hyperparameter Sensitivity} \label{sec:sensitivity}
Here we further analyze the influence of two key hyperparameters with a known percentage of 80\%. For each experimental setting, we adopt NCI as the anomaly detection method and repeat the attack ten times and report the average performance together with its range. All values are rounded to two decimal places, which may cause very small numbers to appear as 0.00.

\begin{figure}[!ht]
    \centering
    \subfloat[AUC]{
    \includegraphics[width=0.49\linewidth]{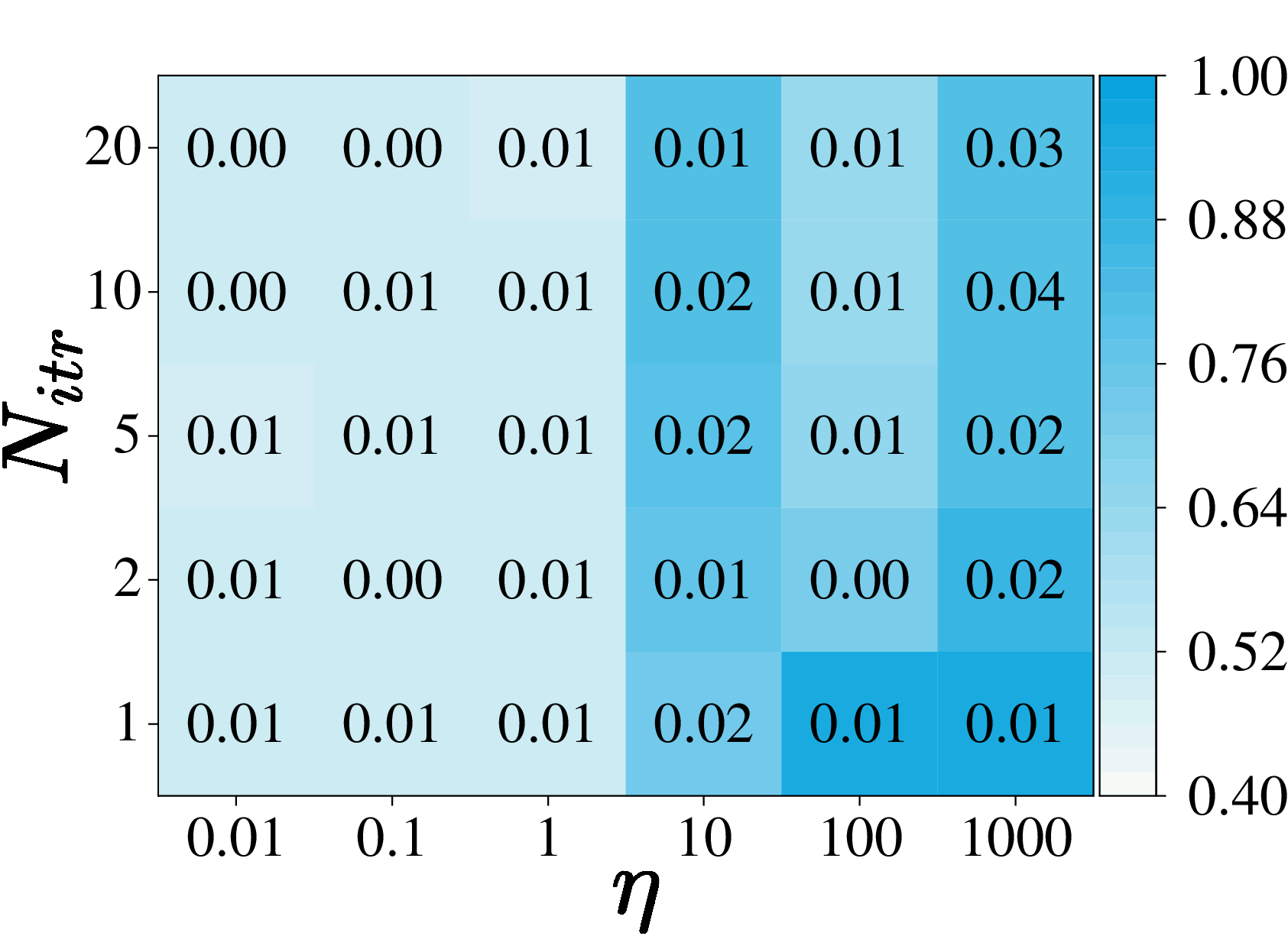}
    }
    \subfloat[TPR@0.1FPR]{
    \includegraphics[width=0.49\linewidth]{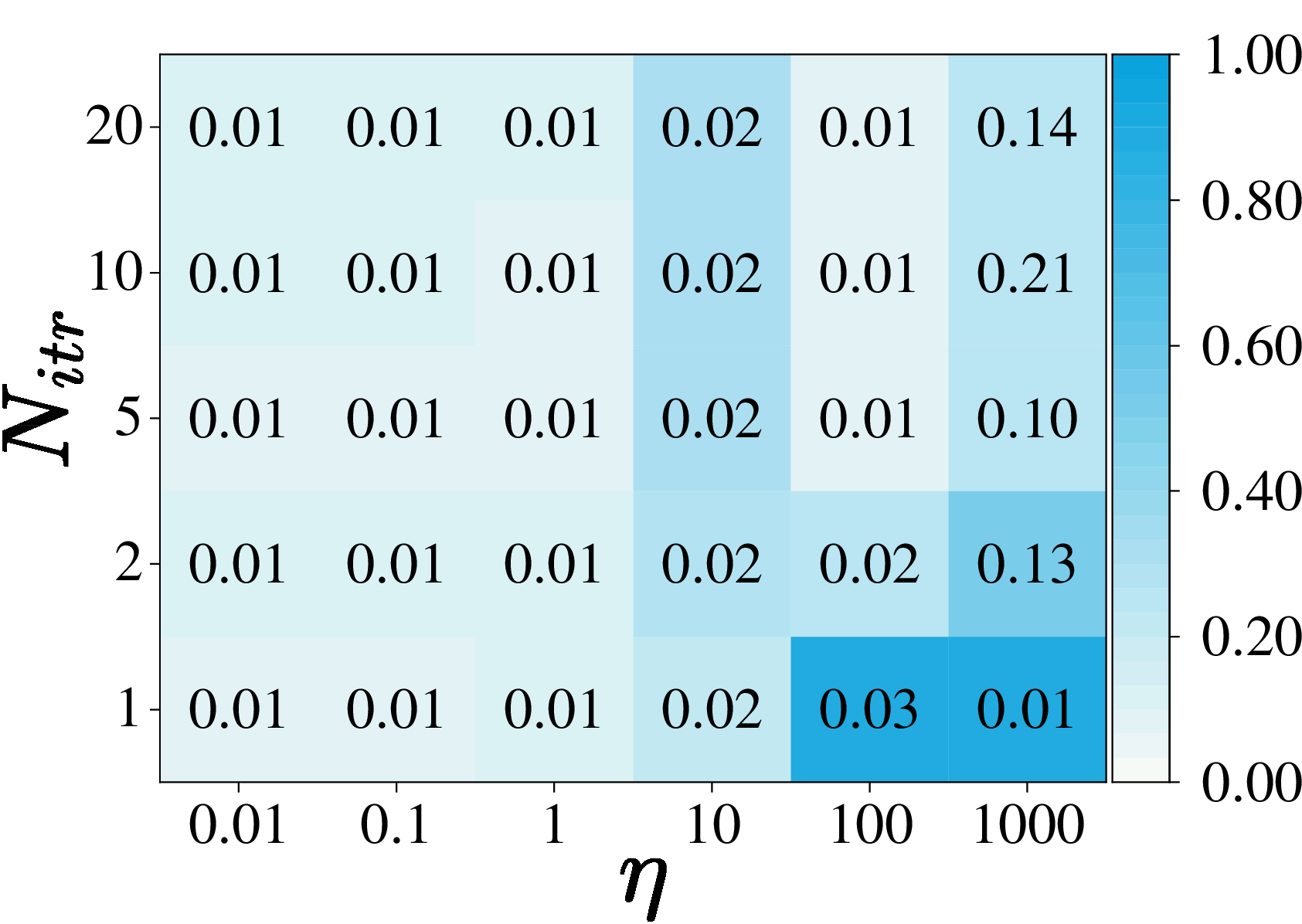}
    }
    \caption{Attack performance on the CIFAR-10 dataset under different parameter settings. Darker blue indicates higher attack performance. The number in each cell denotes half of the range (i.e., half the max–min difference) across ten runs.}
    \label{fig:sen_cifar10}
\end{figure}

\begin{figure}[!ht]
    \centering
    \subfloat[AUC]{
    \includegraphics[width=0.49\linewidth]{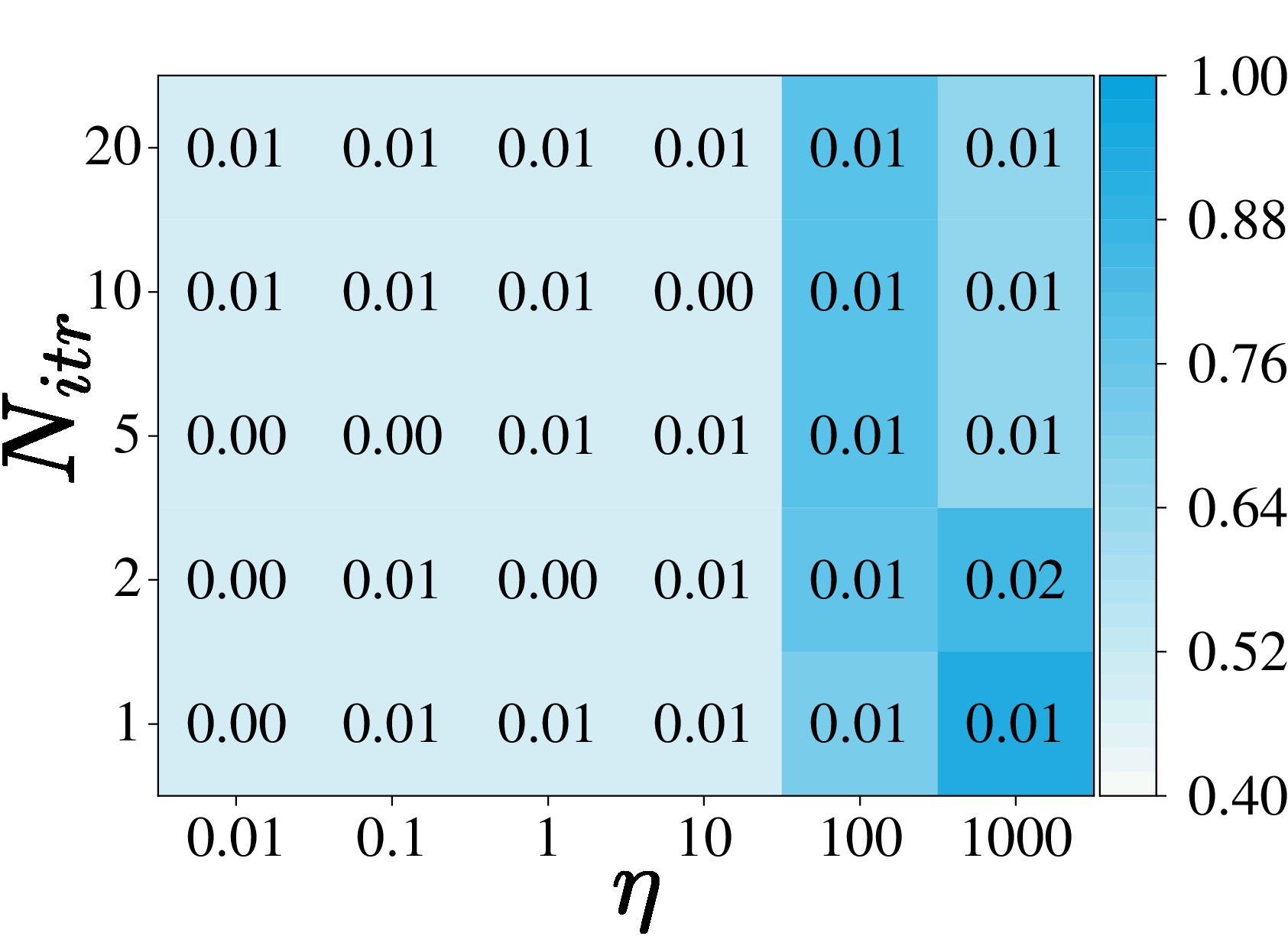}
    }
    \subfloat[TPR@0.1FPR]{
    \includegraphics[width=0.49\linewidth]{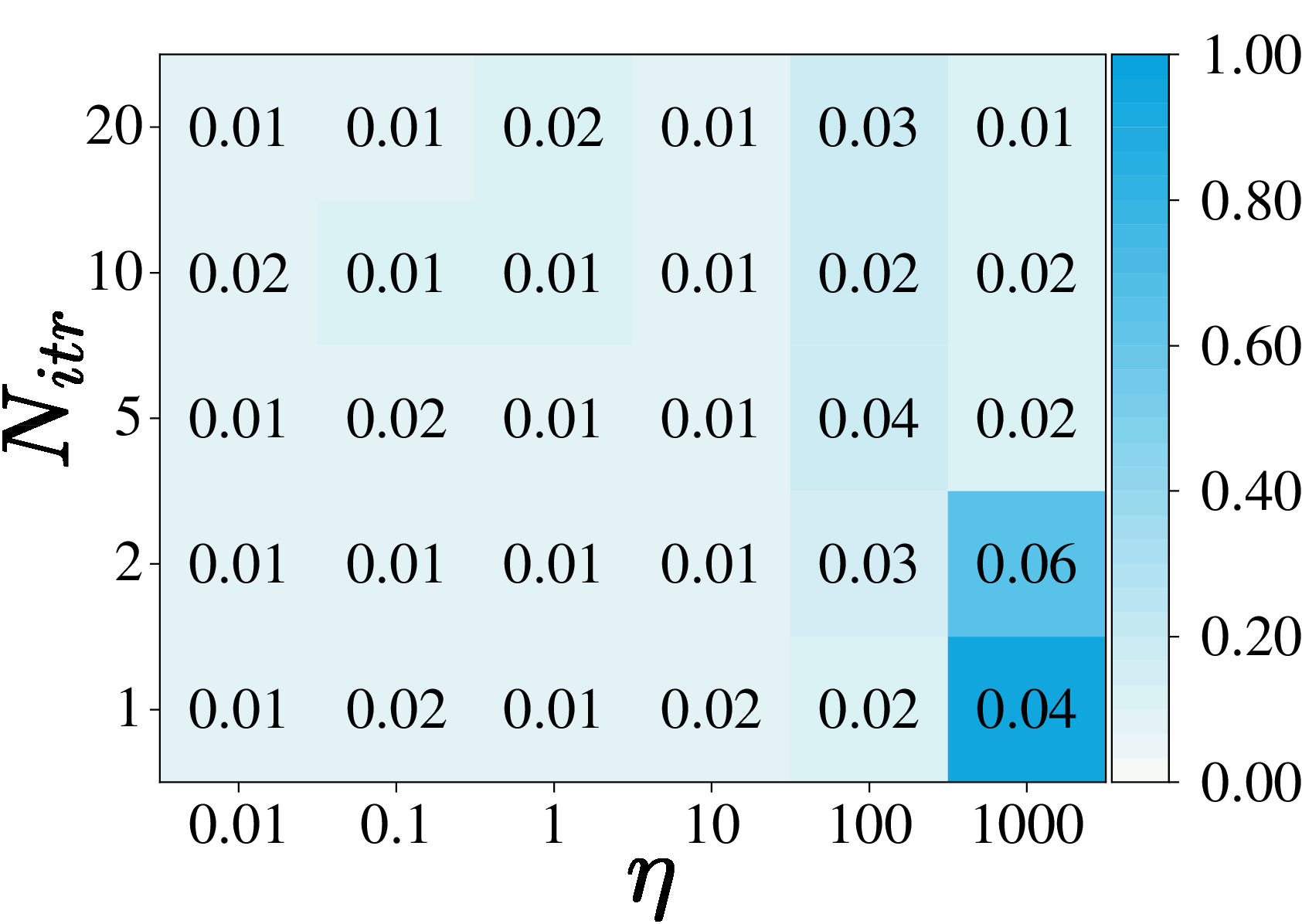}
    }
    \caption{Attack performance on the STL-10 dataset under different parameter settings. Darker blue indicates higher attack performance. The number in each cell denotes half of the range (i.e., half the max–min difference) across ten runs.}
    \label{fig:sen_stl10}
\end{figure}

\begin{figure}[!ht]
    \centering
    \subfloat[AUC]{
    \includegraphics[width=0.49\linewidth]{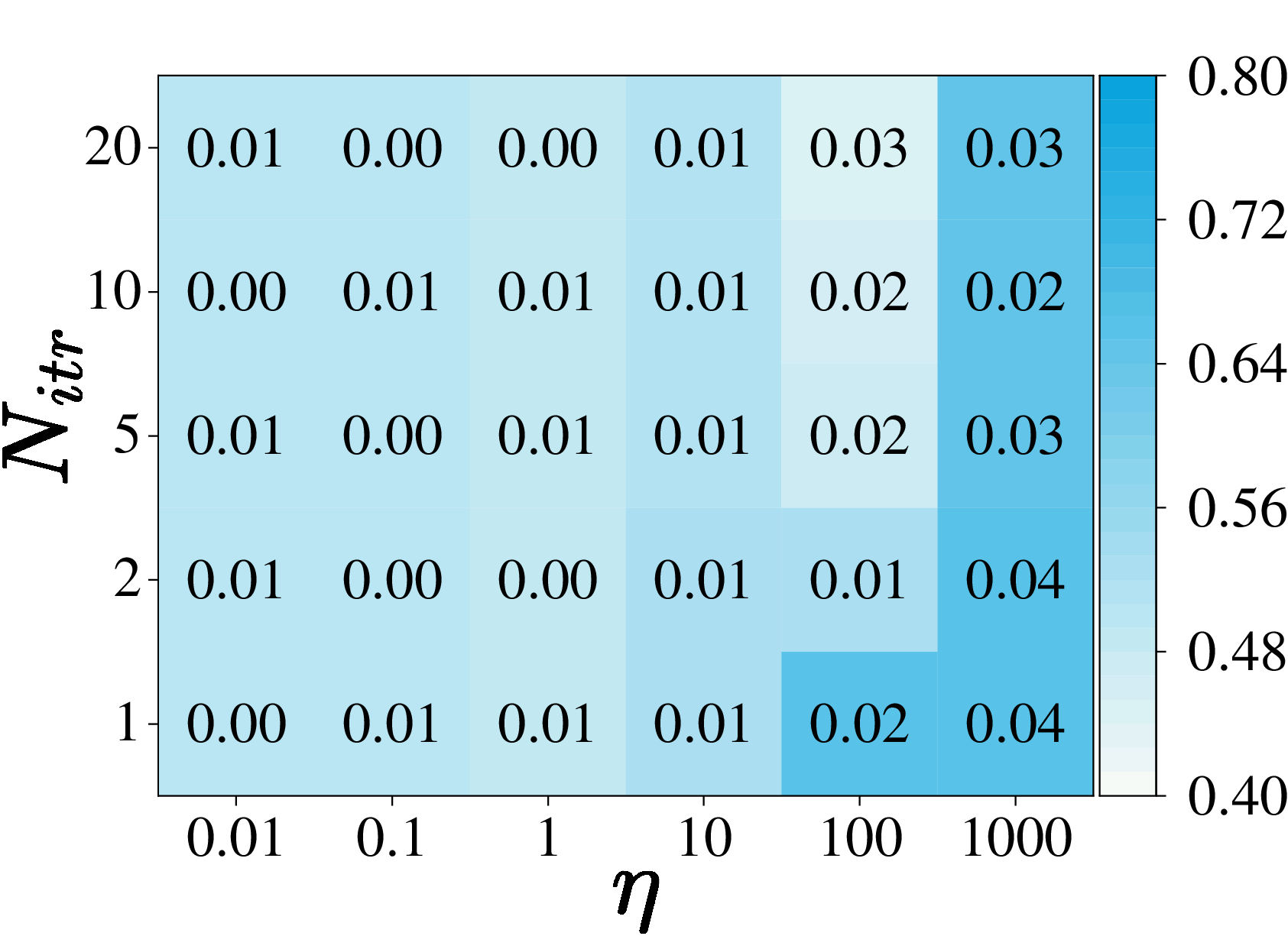}
    }
    \subfloat[TPR@0.1FPR]{
    \includegraphics[width=0.49\linewidth]{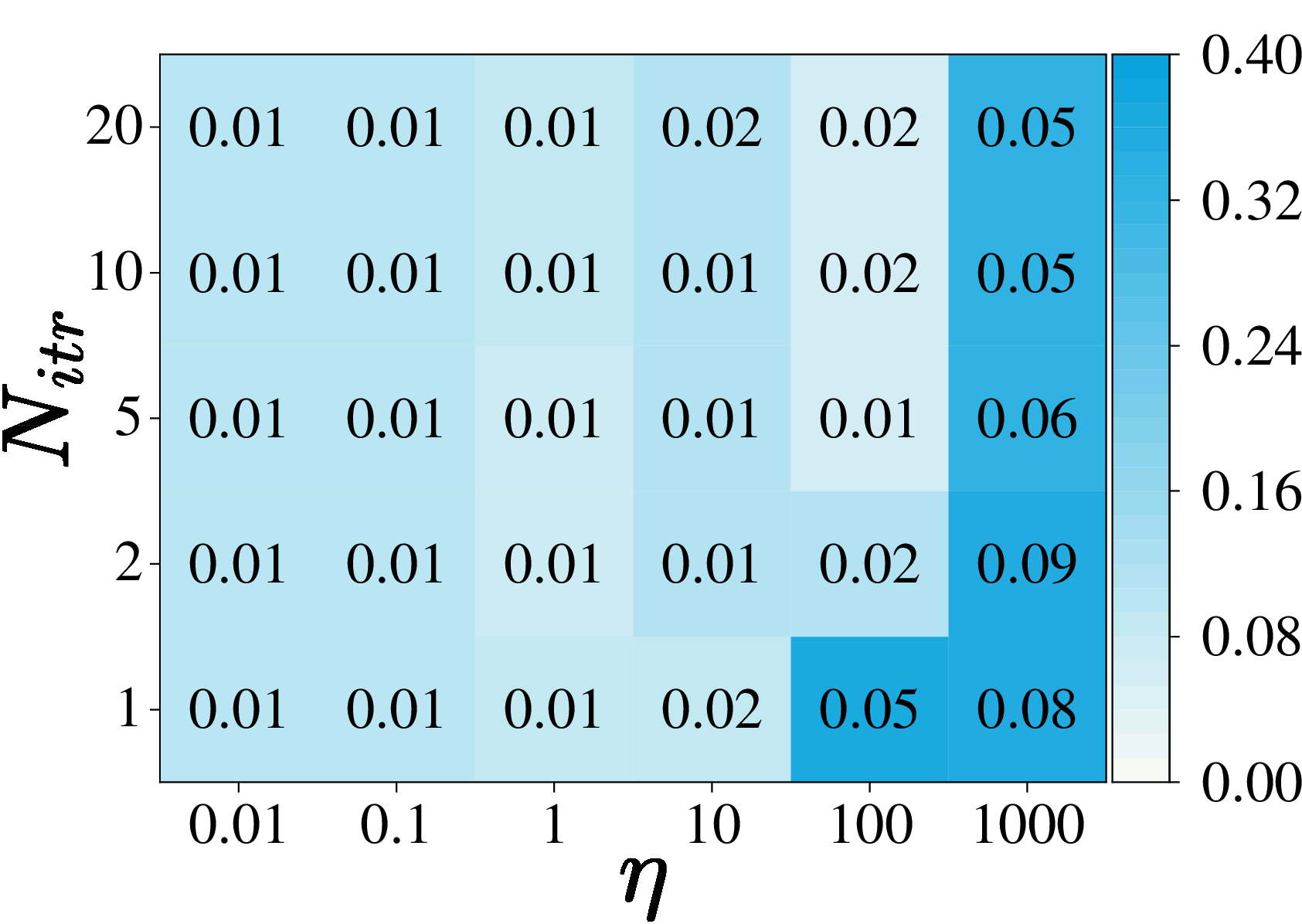}
    }
    \caption{Attack performance on the Fashion-MNIST dataset under different parameter settings. Darker blue indicates higher attack performance. The number in each cell denotes half of the range (i.e., half the max–min difference) across ten runs.}
    \label{fig:sen_fashion}
\end{figure}

\begin{figure}[!ht]
    \centering
    \subfloat[AUC]{
    \includegraphics[width=0.49\linewidth]{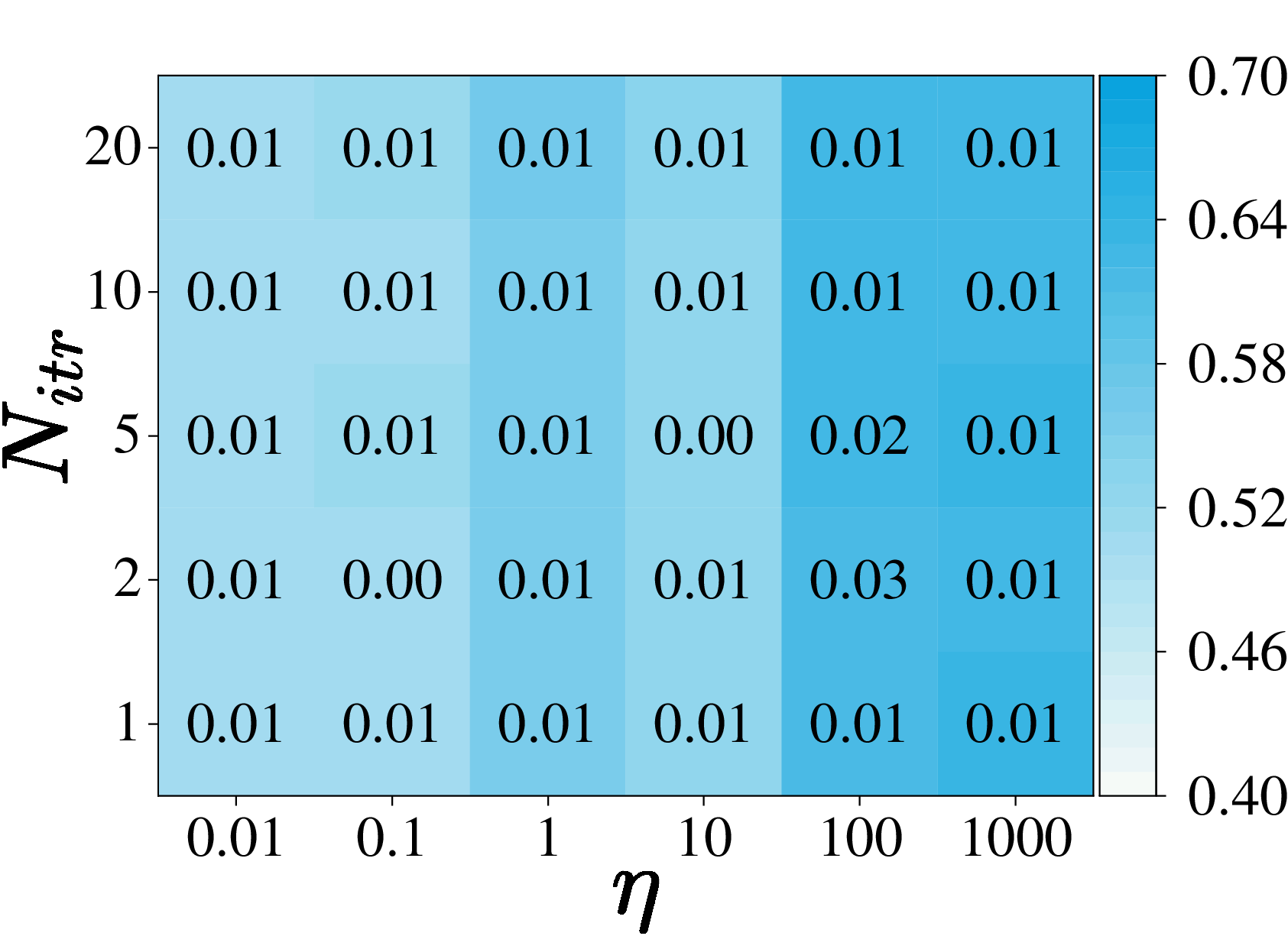}
    }
    \subfloat[TPR@0.1FPR]{
    \includegraphics[width=0.49\linewidth]{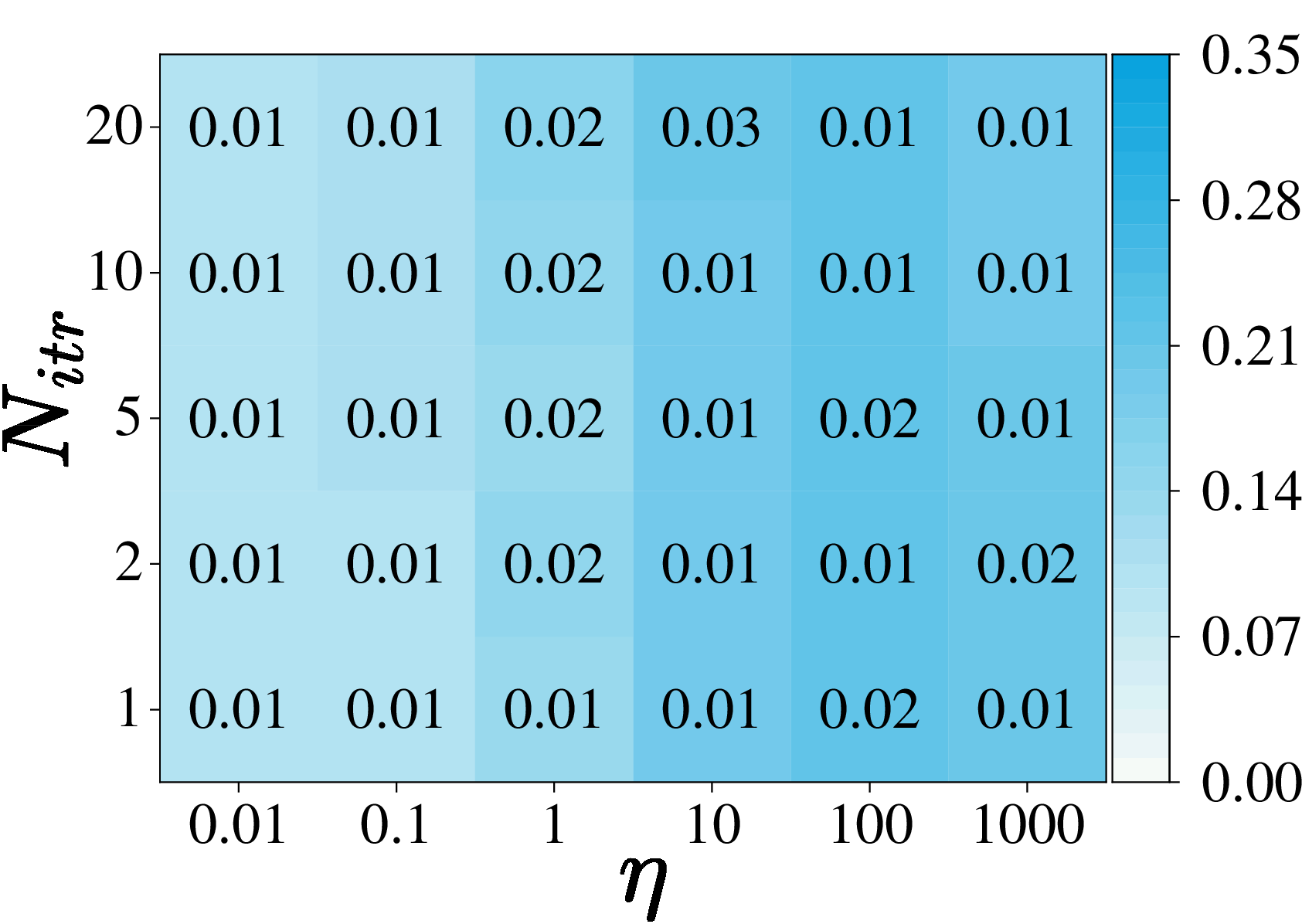}
    }
    \caption{Attack performance on the Epsilon dataset under different parameter settings. Darker blue indicates higher attack performance. The number in each cell denotes half of the range (i.e., half the max–min difference) across ten runs.}
    \label{fig:sen_epsilon}
\end{figure}

Figure \ref{fig:sen_cifar10}, Figure \ref{fig:sen_stl10}, Figuure \ref{fig:sen_fashion} and Figure\ref{fig:sen_epsilon} report the results of applying our attack to four datasets. Based on the experimental results, we identify two parameter settings that can lead to strong attack performance.

The first approach is to use a large step size $\eta$ together with a small number of optimization iterations $N_{itr}$. As shown in the heatmaps, when $N_{itr}$ is fixed, increasing $\eta$ generally improves the attack performance within the range we evaluated. This is because, for a sample that follows the normal data distribution, the loss produced by a well-trained target model remains small even when some features are missing. A larger step size is therefore necessary to amplify the gradient signal and make the optimization updates effective. In addition, when $\eta$ is sufficiently large, only a few optimization iterations are needed to achieve strong results, while excessively increasing $N_{itr}$ may even reduce the attack effectiveness.

The second approach is to choose a moderate step size $\eta$ and perform multiple rounds of optimization (e.g., $N_{itr}=20, \eta=10$ on CIFAR-10 or $N_{itr}=20, \eta=100$ on STL-10). However, this strategy does not outperform the first approach and increases the overall computational cost—especially in black-box settings, where each optimization step requires $(m+1)$ model queries. For this reason, we recommend using a relatively large $\eta$ with only a single optimization iteration, which provides a more efficient and stable configuration.

\subsection{Ablation Study}
\begin{figure*}
    \centering
    \subfloat[CIFAR-10]{
    \includegraphics[width=0.22\linewidth]{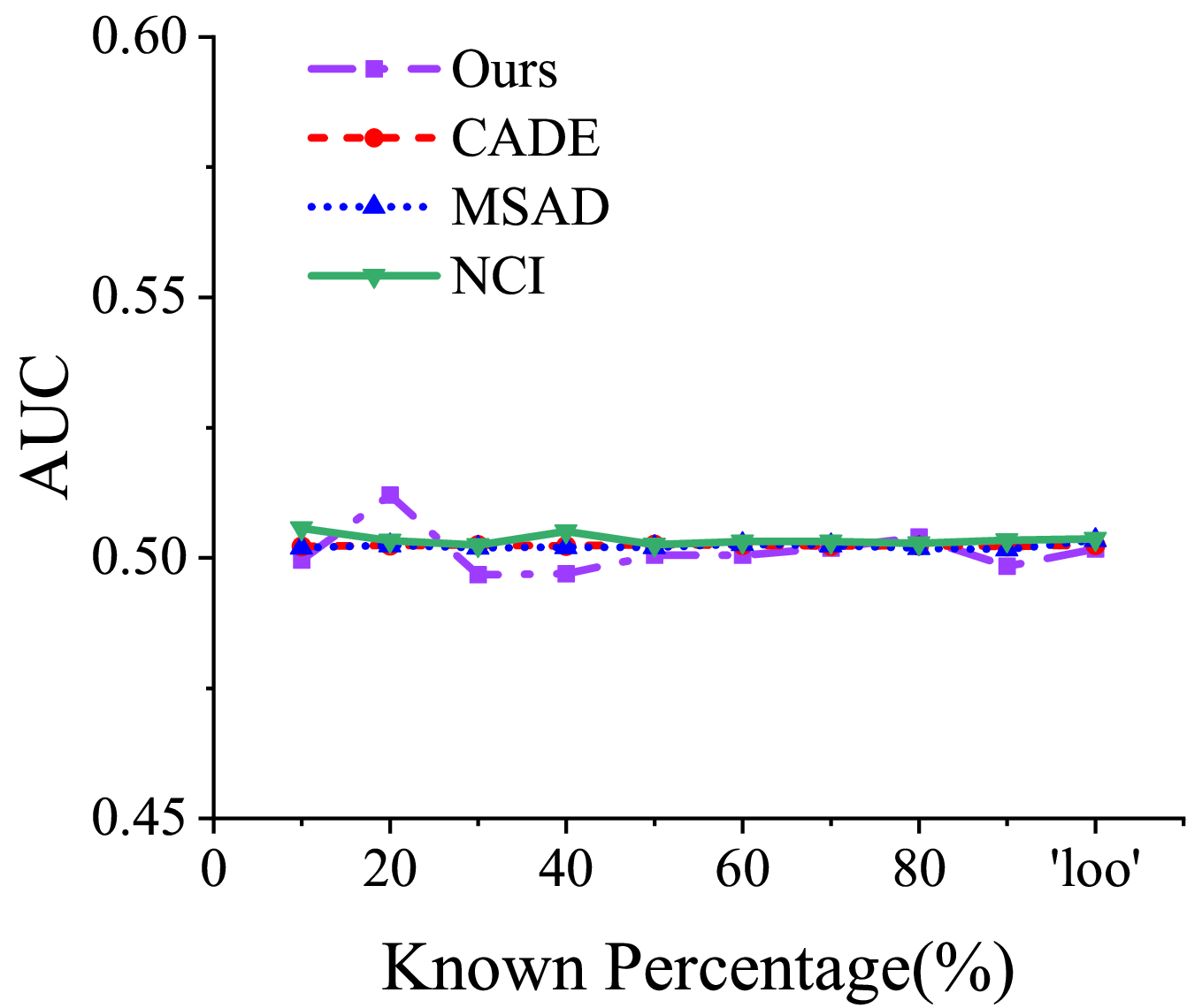}
    }
    \subfloat[STL-10]{
    \includegraphics[width=0.22\linewidth]{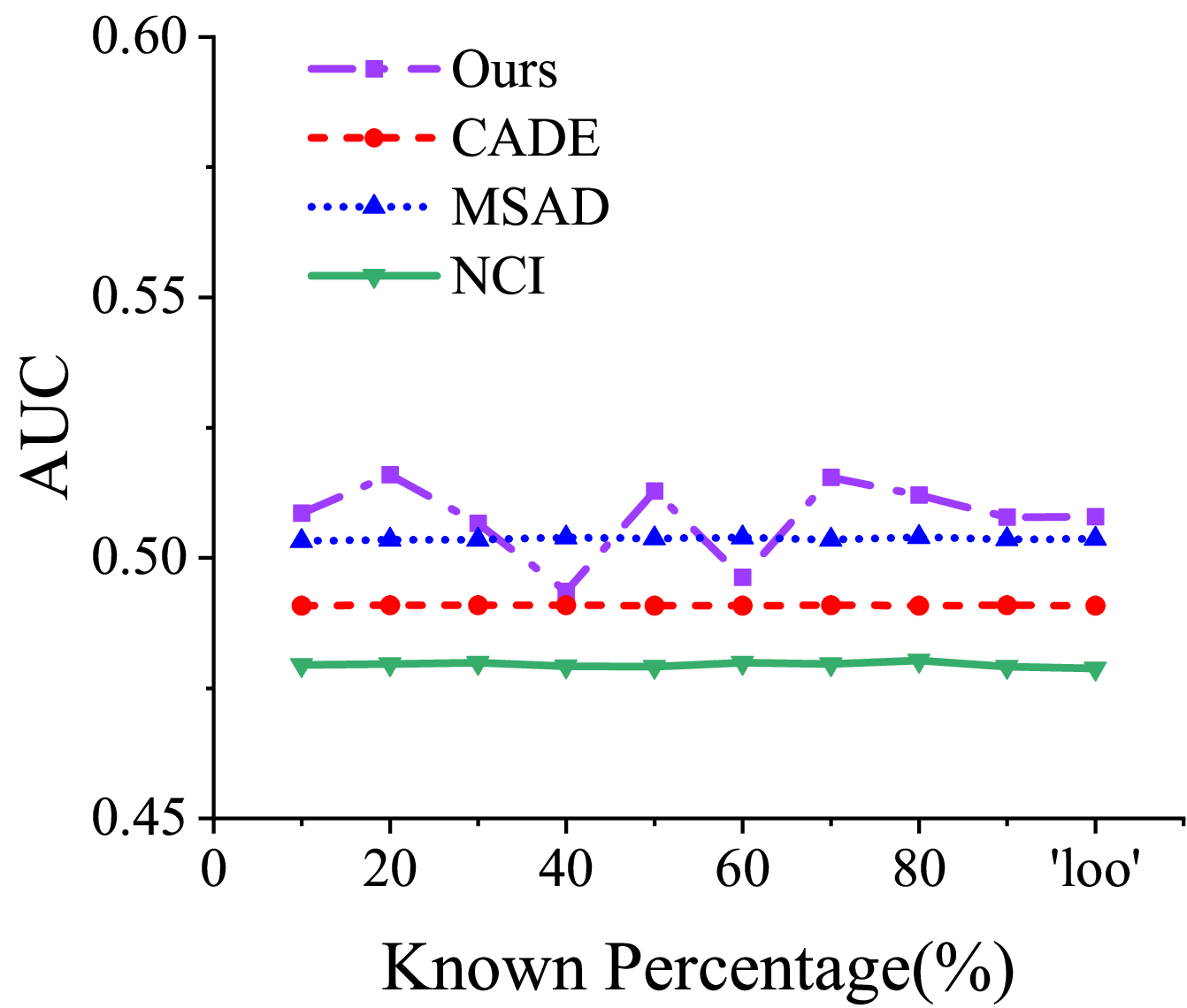}
    }
    \subfloat[Fashion-MNIST]{
    \includegraphics[width=0.22\linewidth]{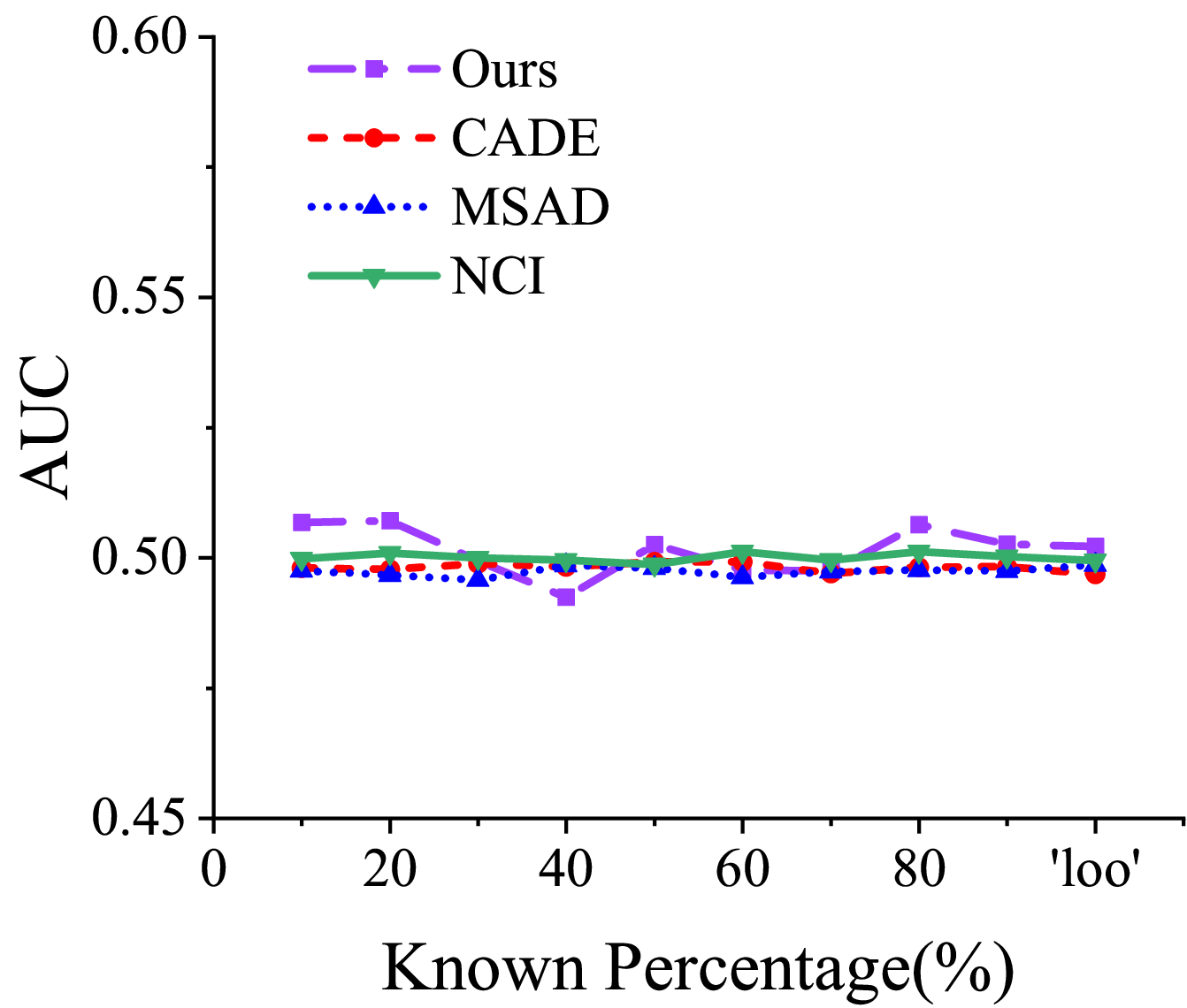}
    }
    \subfloat[Epsilon]{
    \includegraphics[width=0.22\linewidth]{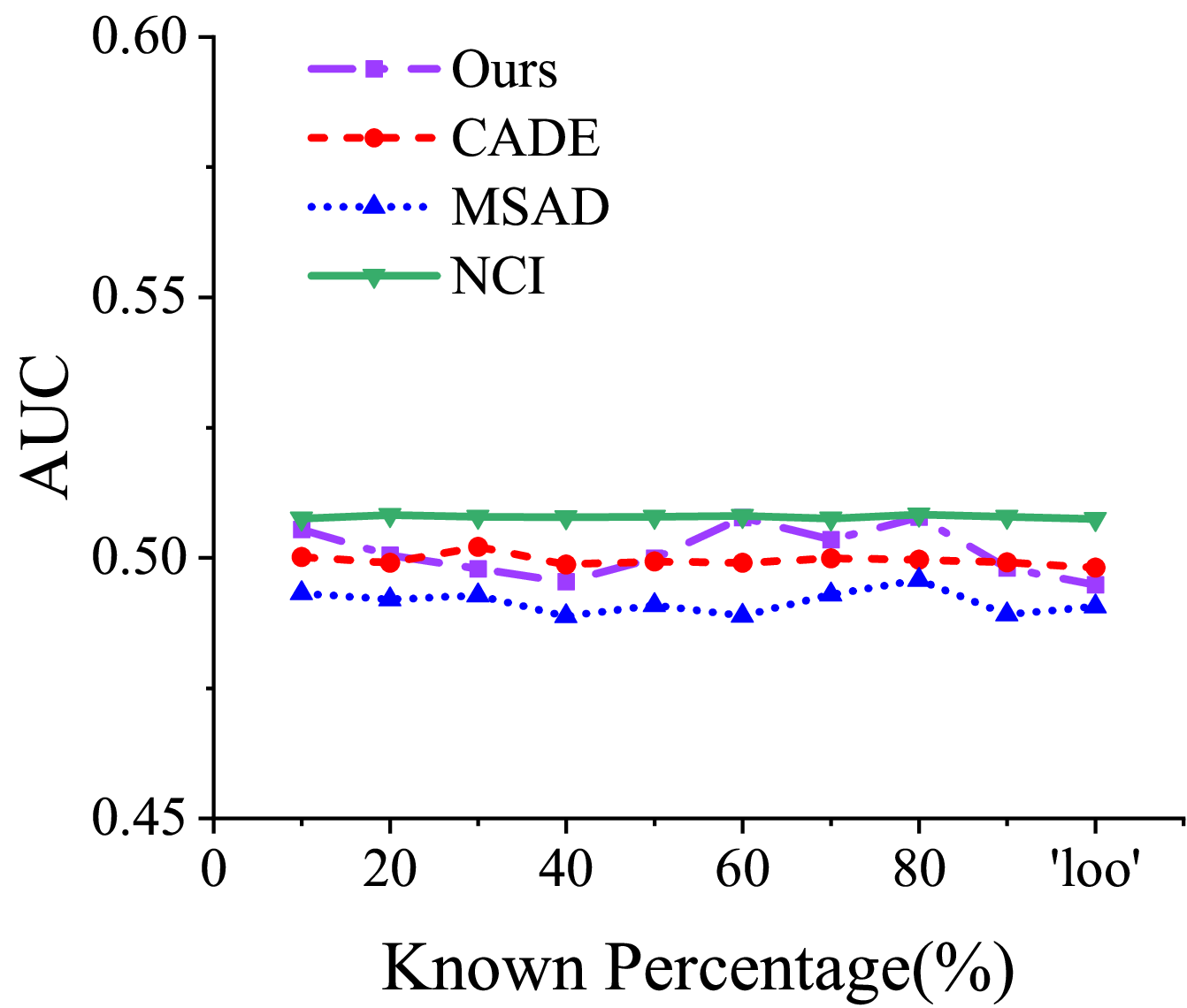}
    }
    \caption{ AUC results when the unknown features are randomly initialized without optimization. We denote the case where only one feature is unknown as “loo”.}
    \label{fig:ablation_auc}
\end{figure*}

\begin{figure*}
    \centering
    \subfloat[CIFAR-10]{
    \includegraphics[width=0.22\linewidth]{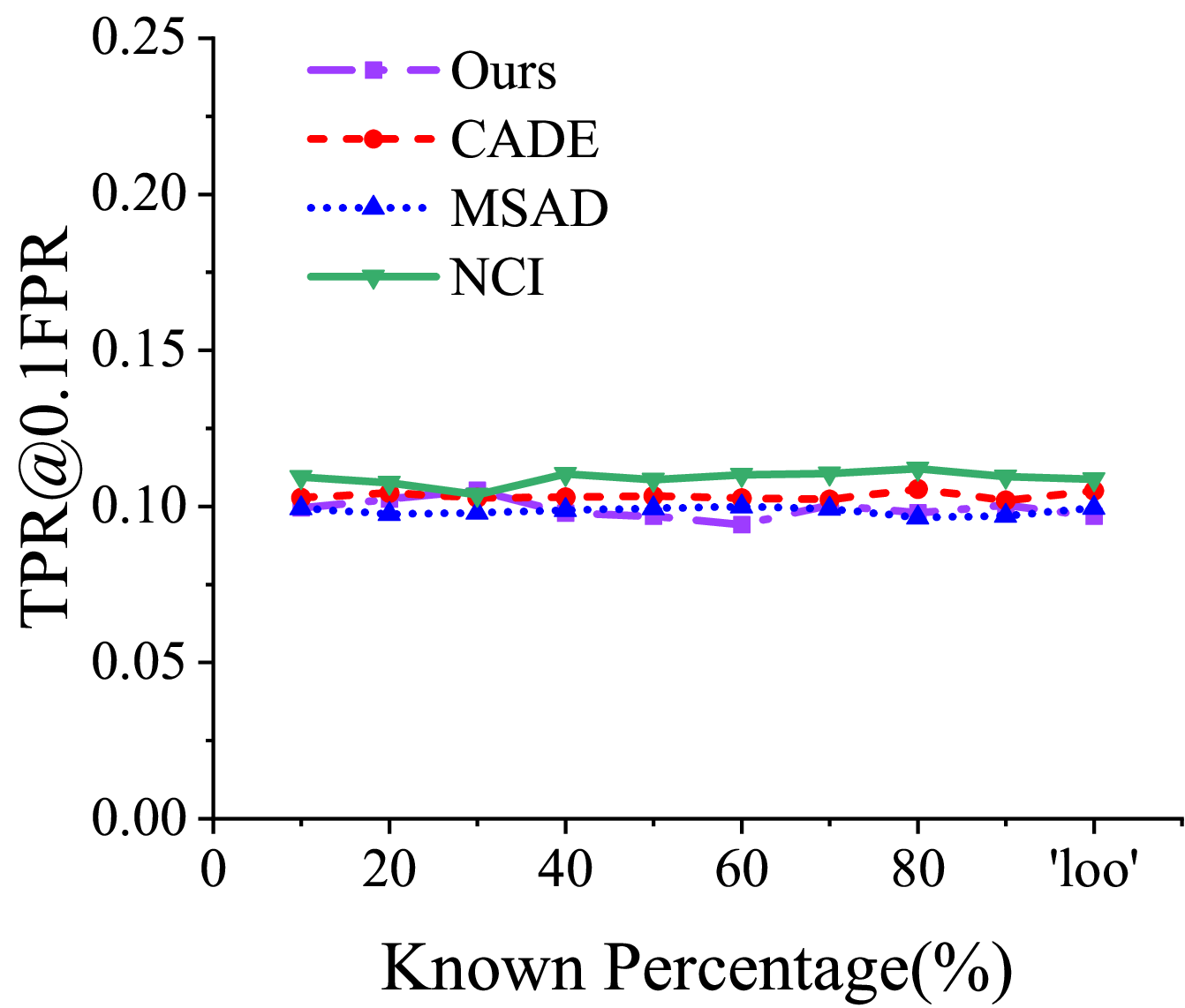}
    }
    \subfloat[STL-10]{
    \includegraphics[width=0.22\linewidth]{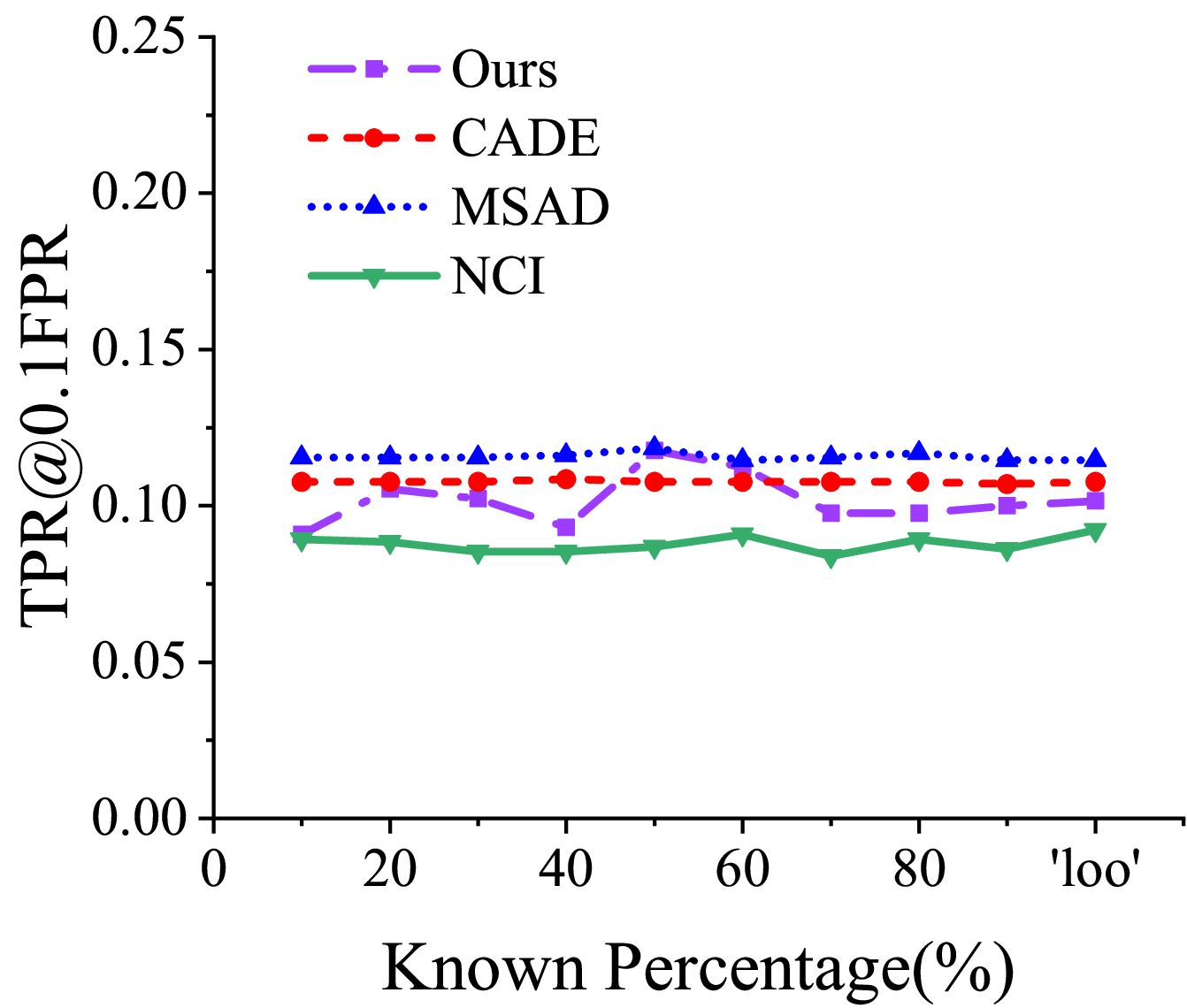}
    }
    \subfloat[Fashion-MNIST]{
    \includegraphics[width=0.22\linewidth]{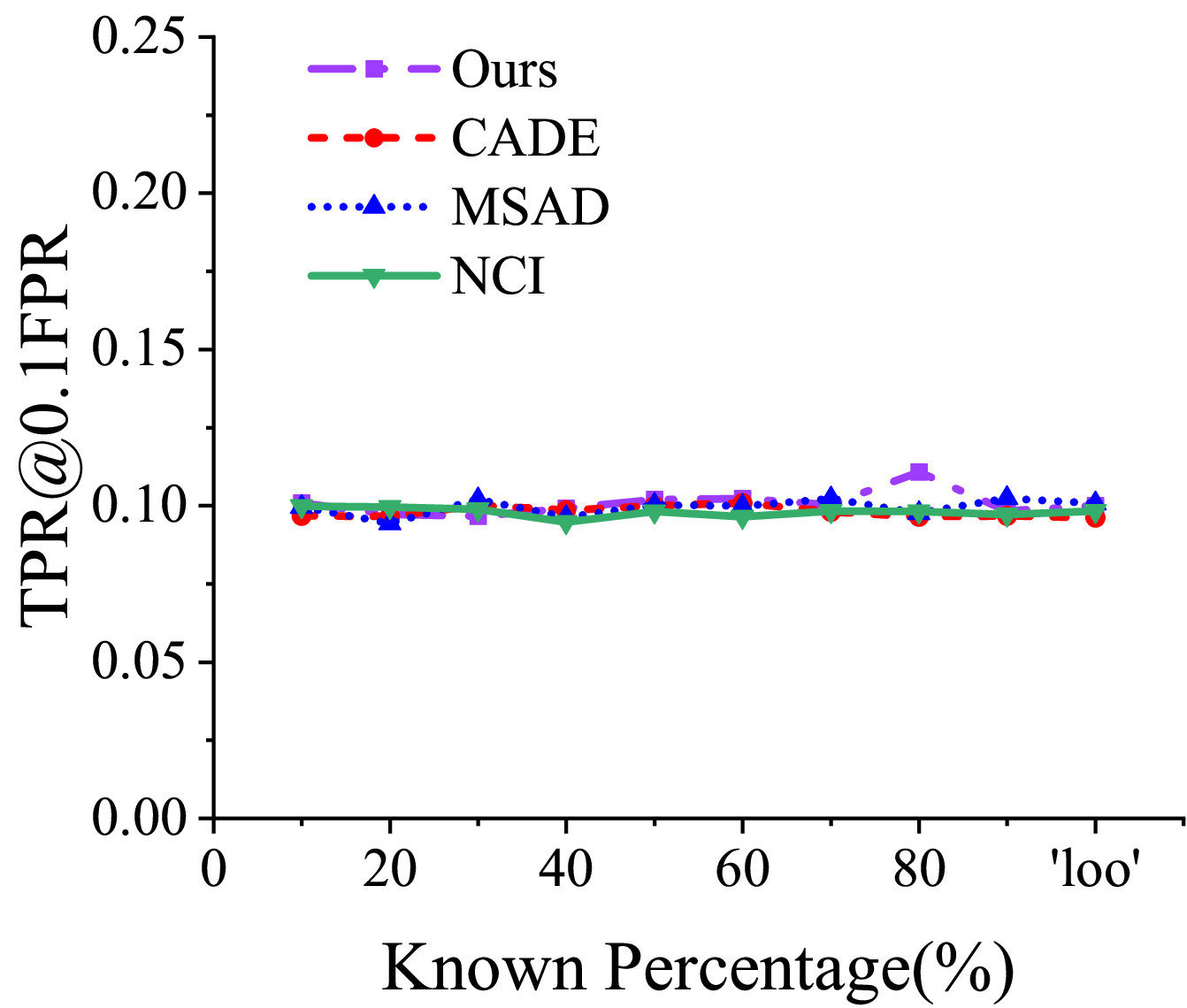}
    }
    \subfloat[Epsilon]{
    \includegraphics[width=0.22\linewidth]{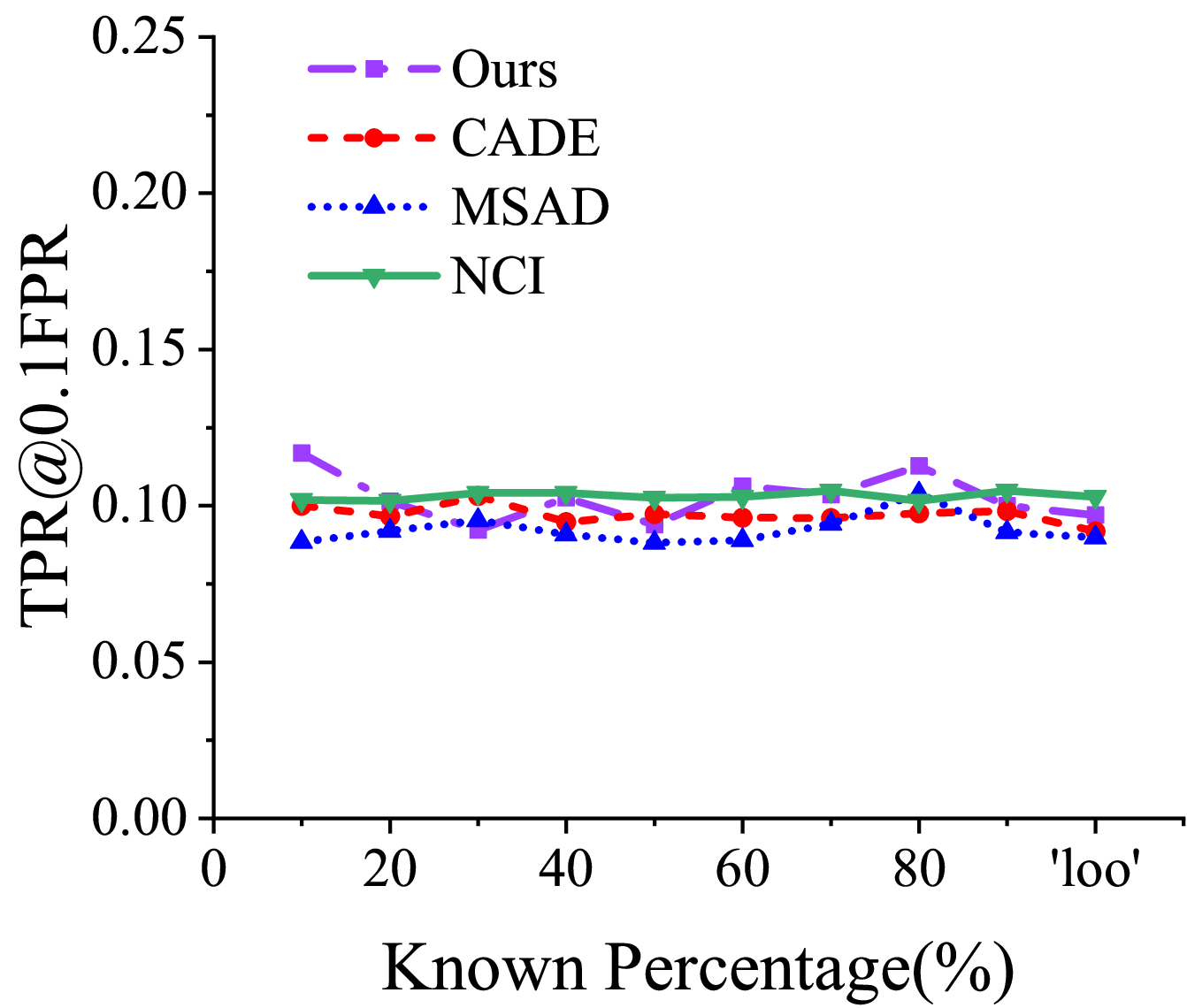}
    }
    \caption{ TPR@0.1FPR results when the unknown features are randomly initialized without optimization. We denote the case where only one feature is unknown as “loo”.}
    \label{fig:ablation_tpr}
\end{figure*}

To further validate the effectiveness of our algorithm, we skip the first stage and directly perform anomaly detection using random initialized unknown features. The results are shown in Figure \ref{fig:ablation_auc} and Figure \ref{fig:ablation_tpr}. As observed, when the unknown features are not optimized through loss minimization, the anomaly detection methods fail to discriminate between reconstructed samples containing member features and those containing non‑member features. This outcome is intuitive and further highlights the critical role of the first-stage reconstruction in enabling successful PFMI.

\subsection{Impact of Feature Importance on Attack Performance}\label{sec:importance}
Since models perform inference based on feature values and different features contribute unequally to the model output, we investigate how the importance of known features influences attack performance. In this section, we use SHAP \cite{NIPS2017_7062} to quantify the contribution of each feature to the prediction of the target model. SHAP is a widely used method for interpreting matchine learning models by assigning an importance value to each feature. It is based on the concept of Shapley value from game theory. Features that contribute more to the output are assigned higher importance values. We evaluate the attack performance under different known feature proportions, using two selection strategies:
\begin{itemize}
    \item K=IF (Keep Important Features as Known Features): The known features are chosen by ranking all features according to their importance scores and selecting the top ones in descending order.
    \item K=UF (Keep Unimportant Features as Unknown Features): The known features are selected as those with the lowest importance scores, based on an ascending ranking of all features.
\end{itemize}

\begin{figure*}
    \centering
    \subfloat[CIFAR-10]{
    \includegraphics[width=0.22\linewidth]{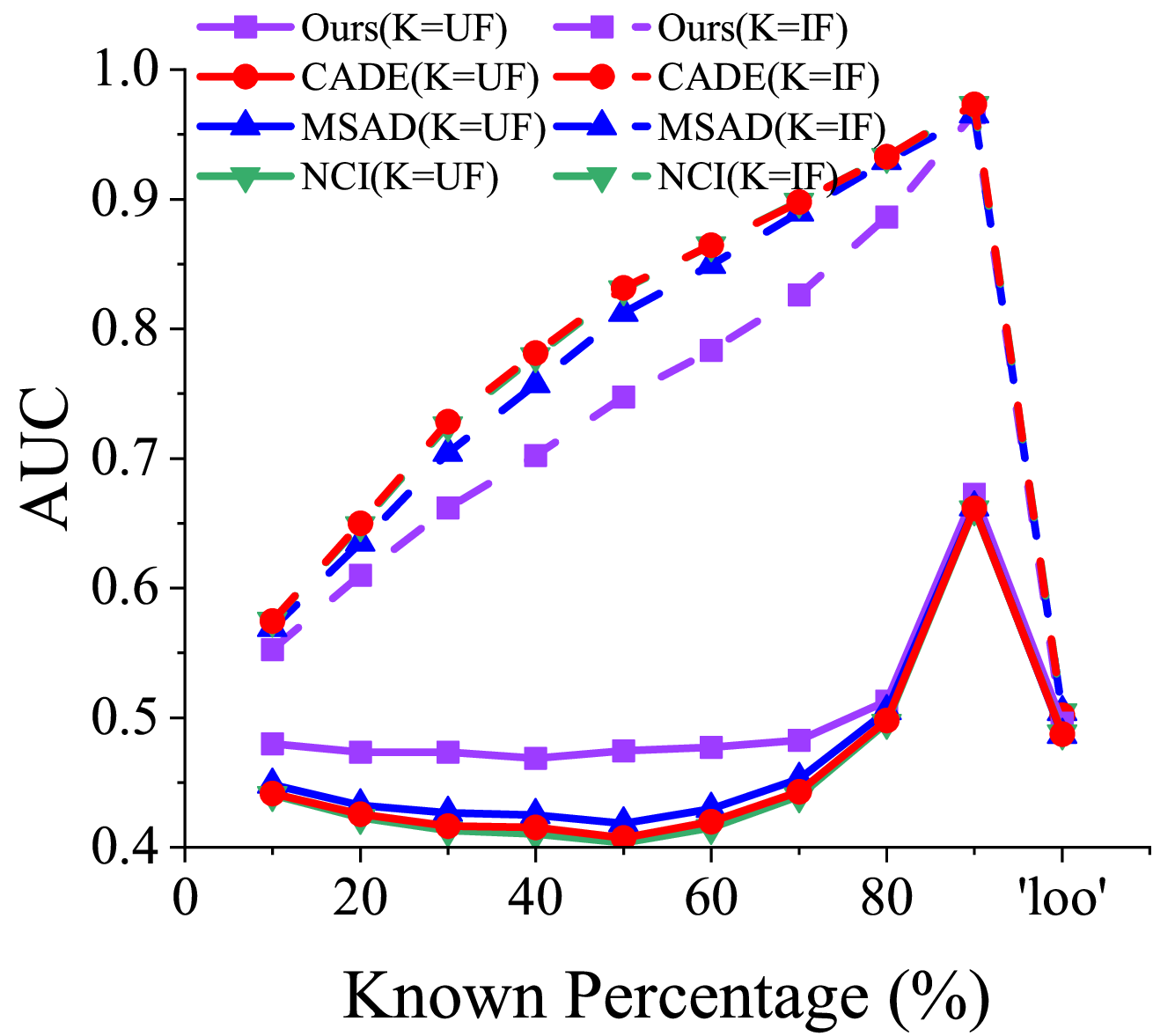}
    }
    \subfloat[STL-10]{
    \includegraphics[width=0.22\linewidth]{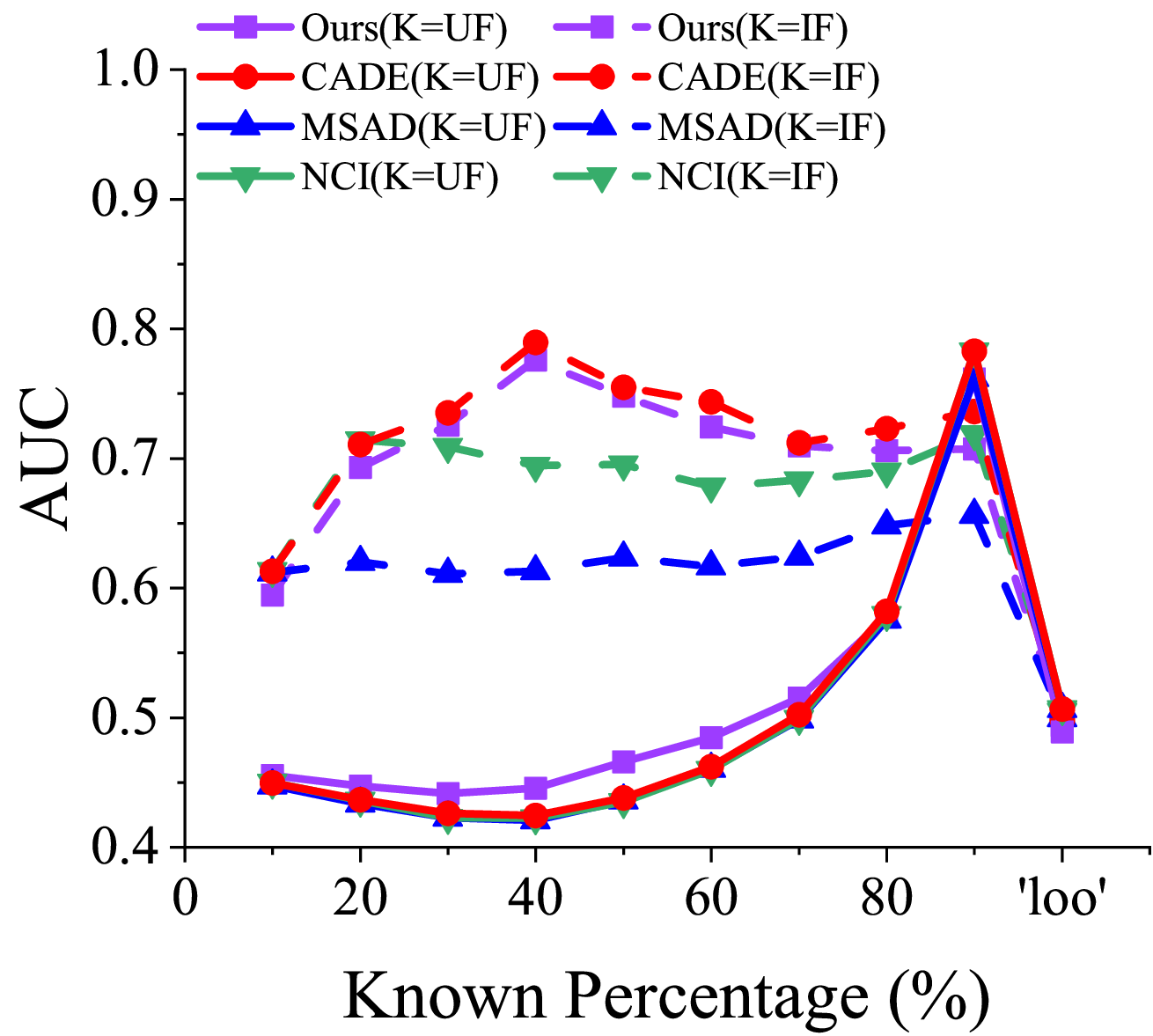}
    }
    \subfloat[Fashion-MNIST]{
    \includegraphics[width=0.22\linewidth]{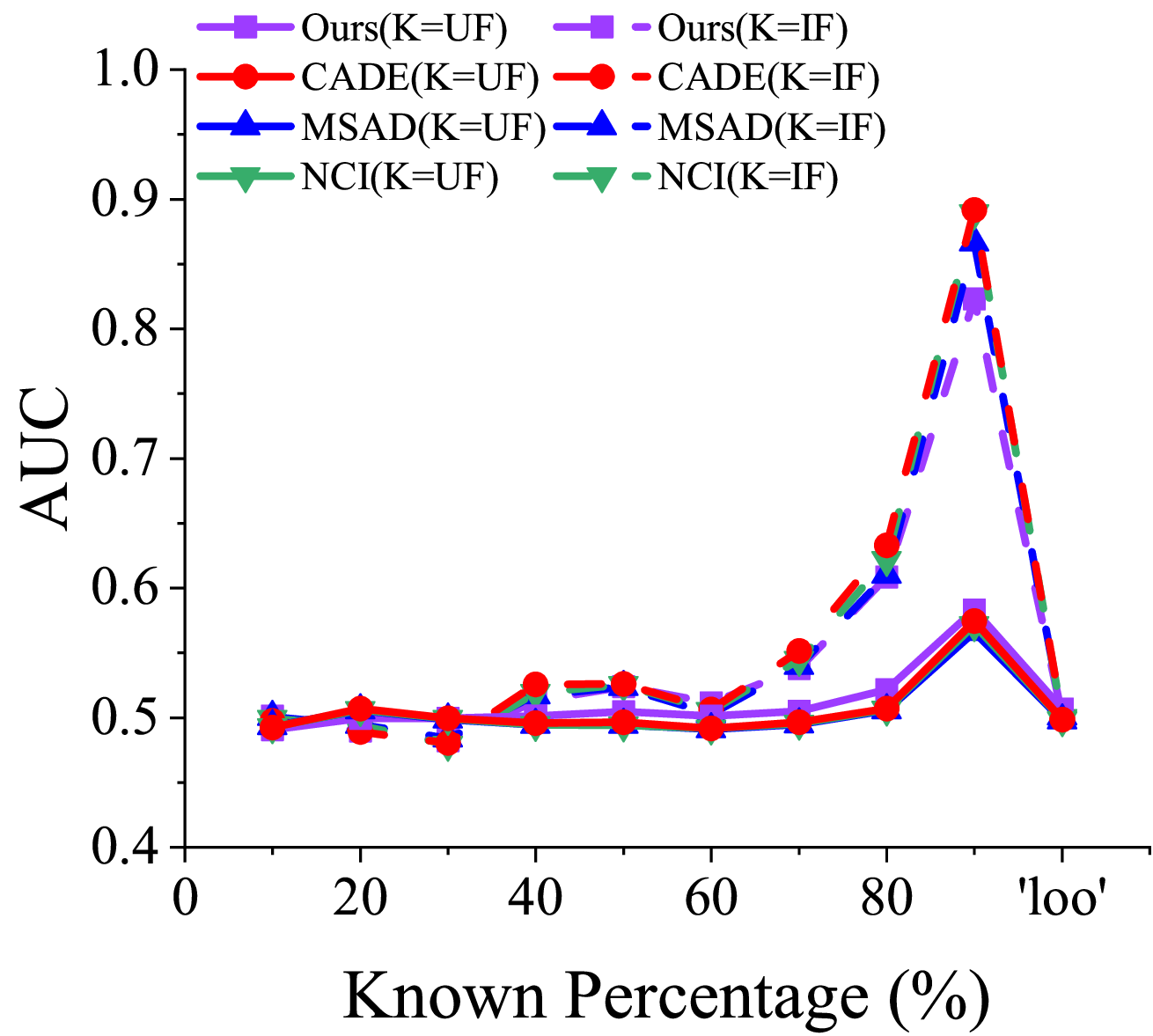}
    }
    \subfloat[Epsilon]{
    \includegraphics[width=0.22\linewidth]{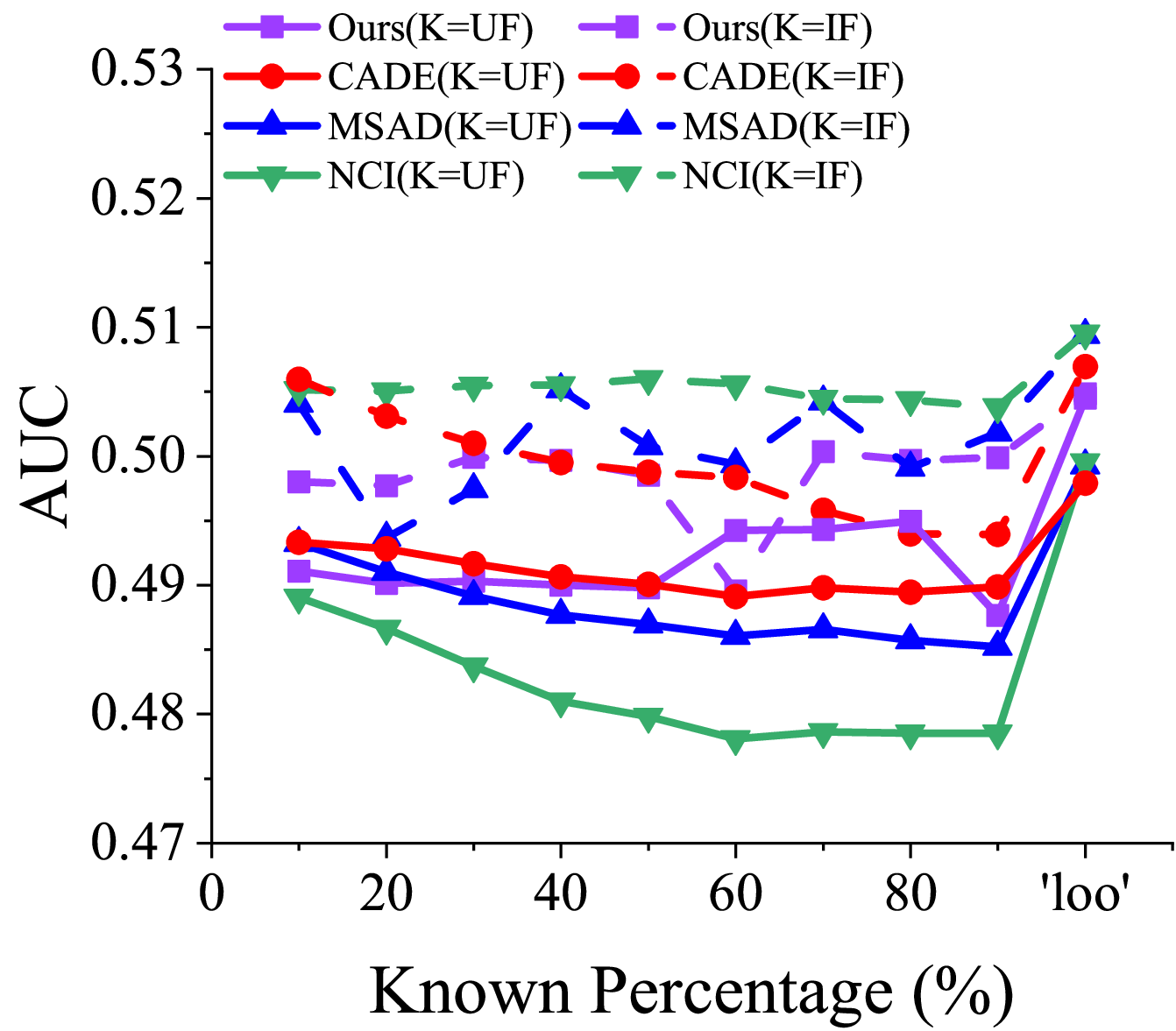}
    }
    \caption{Impact of known feature importance on attack performance (AUC) across different datasets. Dashed lines represent scenarios where known features are the most important ones (K=IF), while solid lines indicate scenarios where the least important features are known (K=UF). We denote the case where only one feature is unknown as “loo”.}
    \label{fig:shap_auc}
\end{figure*}

The results are presented in Figure \ref{fig:shap_auc}. Several observations emerge:
\begin{itemize}
    \item When the known features align with those that have high SHAP importance, the attack performs much better than when only low-importance features are available. This is because important features have a strong influence on the model’s predictions, whereas unimportant features contribute very little. As a result, those unimportant features provide only weak gradient signals and offer less meaningful guidance for the optimization process of the first stage.

    \item By comparing Figure \ref{fig:auc_nf} and Figure \ref{fig:shap_auc}, we observe that when a large portion of the features is known, using only high-importance features as the known ones can actually lead to worse performance than selecting the known features at random, particularly for STL-10 and Epsilon. This happens because once most important features are already known, the remaining unknown features are largely non-essential and contribute very little to the loss. In other words, the values of these remaining features can vary almost arbitrarily without meaningfully affecting the model’s prediction, leaving the attack with too little membership-relevant signal to exploit.
    
\end{itemize}

These results offer insights for both launching and defending against such attacks. For attackers, achieving strong performance requires keeping the number of important known features within a reasonable range—having too few or too many can both lead to degraded attack effectiveness. For defenders, instead of protecting a large number of features, it may be more effective to prioritize the protection of high-importance features, as these features contribute most to the  memorization behavior of the model.

\subsection{Black-box Attack}\label{exp:black_box}
\begin{figure*}
    \centering
    \subfloat[CIFAR-10]{
    \includegraphics[width=0.22\linewidth]{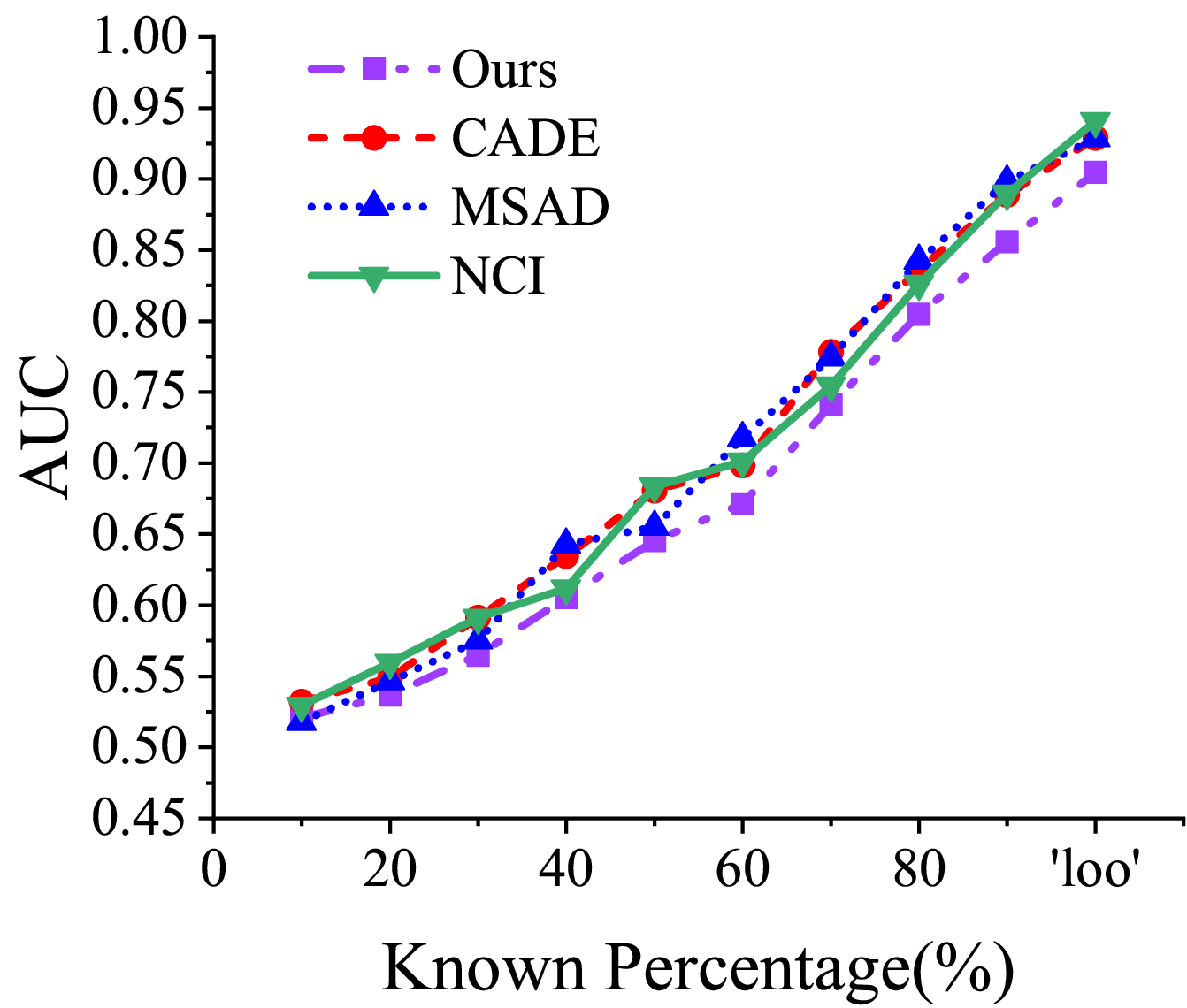}
    }
    \subfloat[STL-10]{
    \includegraphics[width=0.22\linewidth]{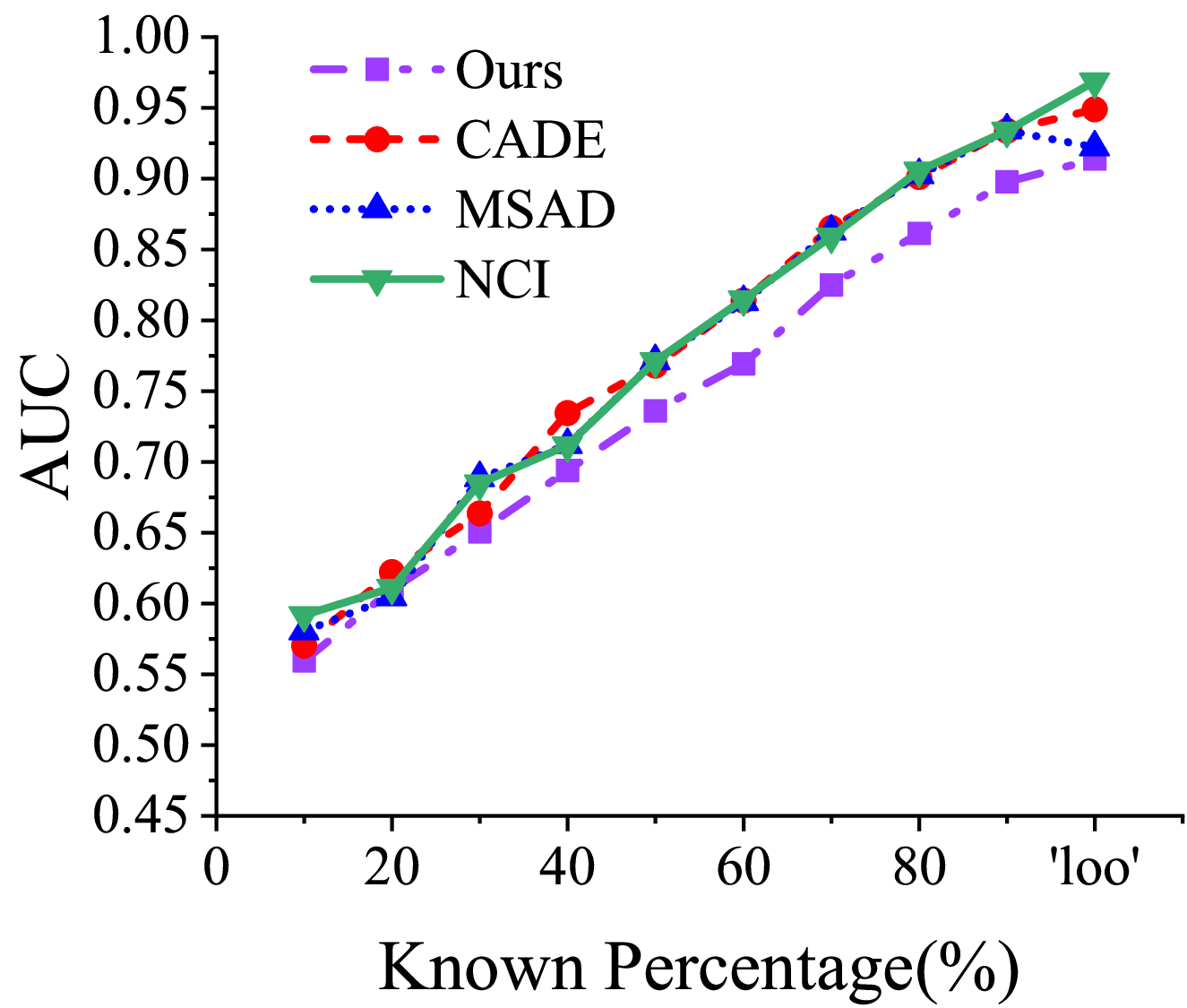}
    }
    \subfloat[Fashion-MNIST]{
    \includegraphics[width=0.22\linewidth]{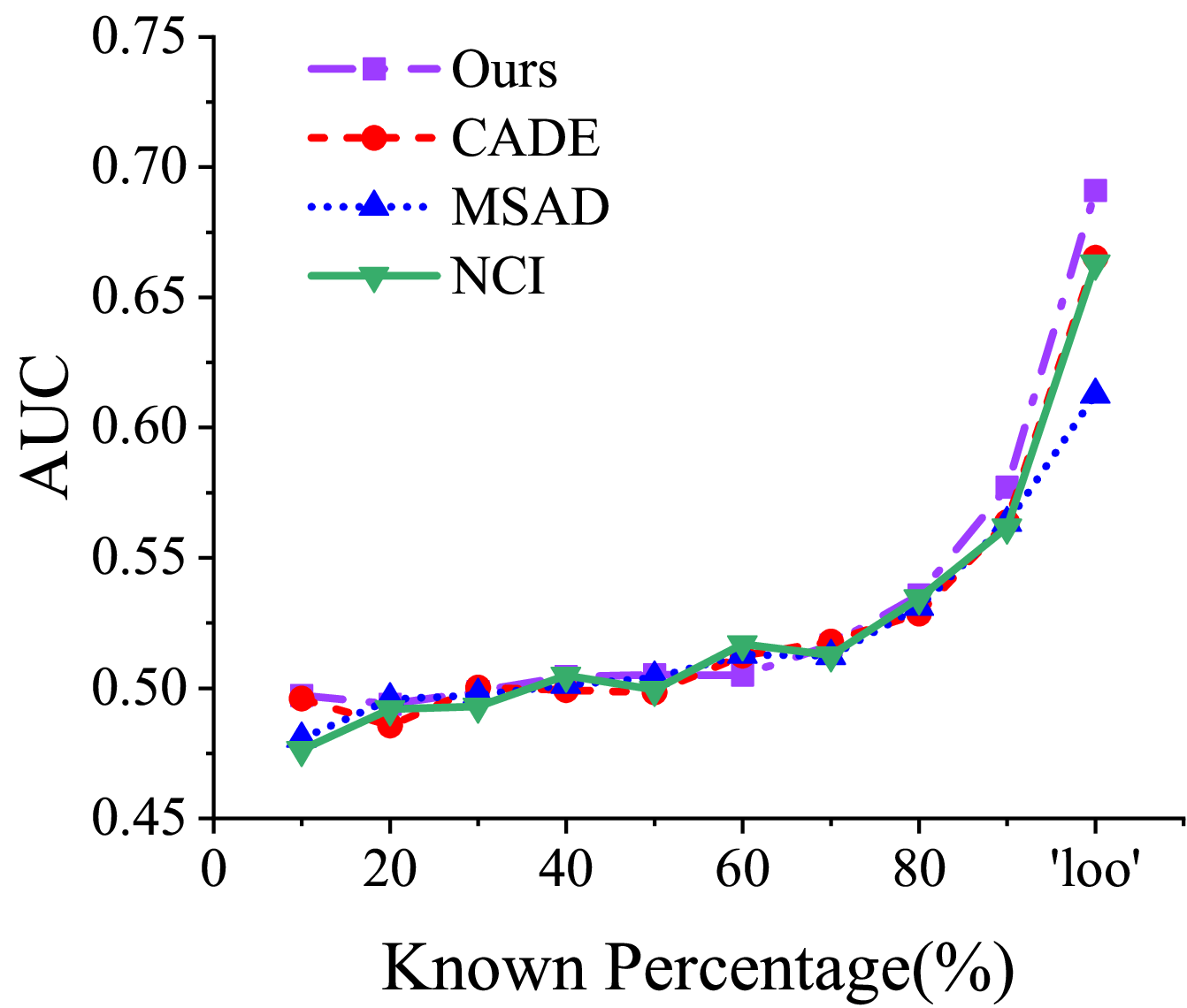}
    }
    \subfloat[Epsilon]{
    \includegraphics[width=0.22\linewidth]{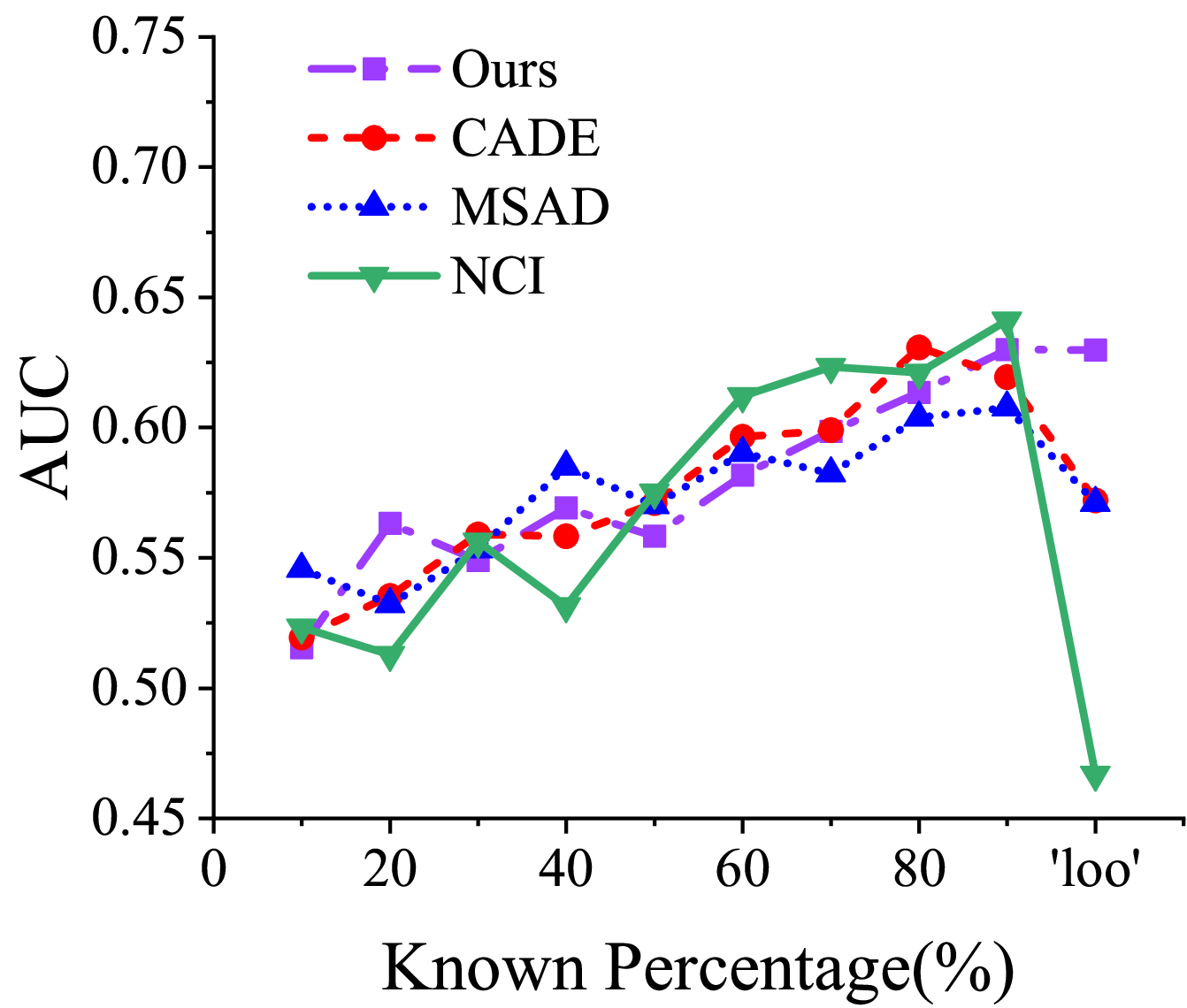}
    }
    \caption{Black-box attack performance (AUC) under varying known feature proportions. The x-axis represents the percentage of known features, and the y-axis shows the corresponding attack AUC. Each curve corresponds to one of the four integrated anomaly detection methods within our attack framework. We denote the case where only one feature is unknown as “loo”.}
    \label{fig:b_auc}
\end{figure*}

\begin{figure*}
    \centering
    \subfloat[CIFAR-10]{
    \includegraphics[width=0.22\linewidth]{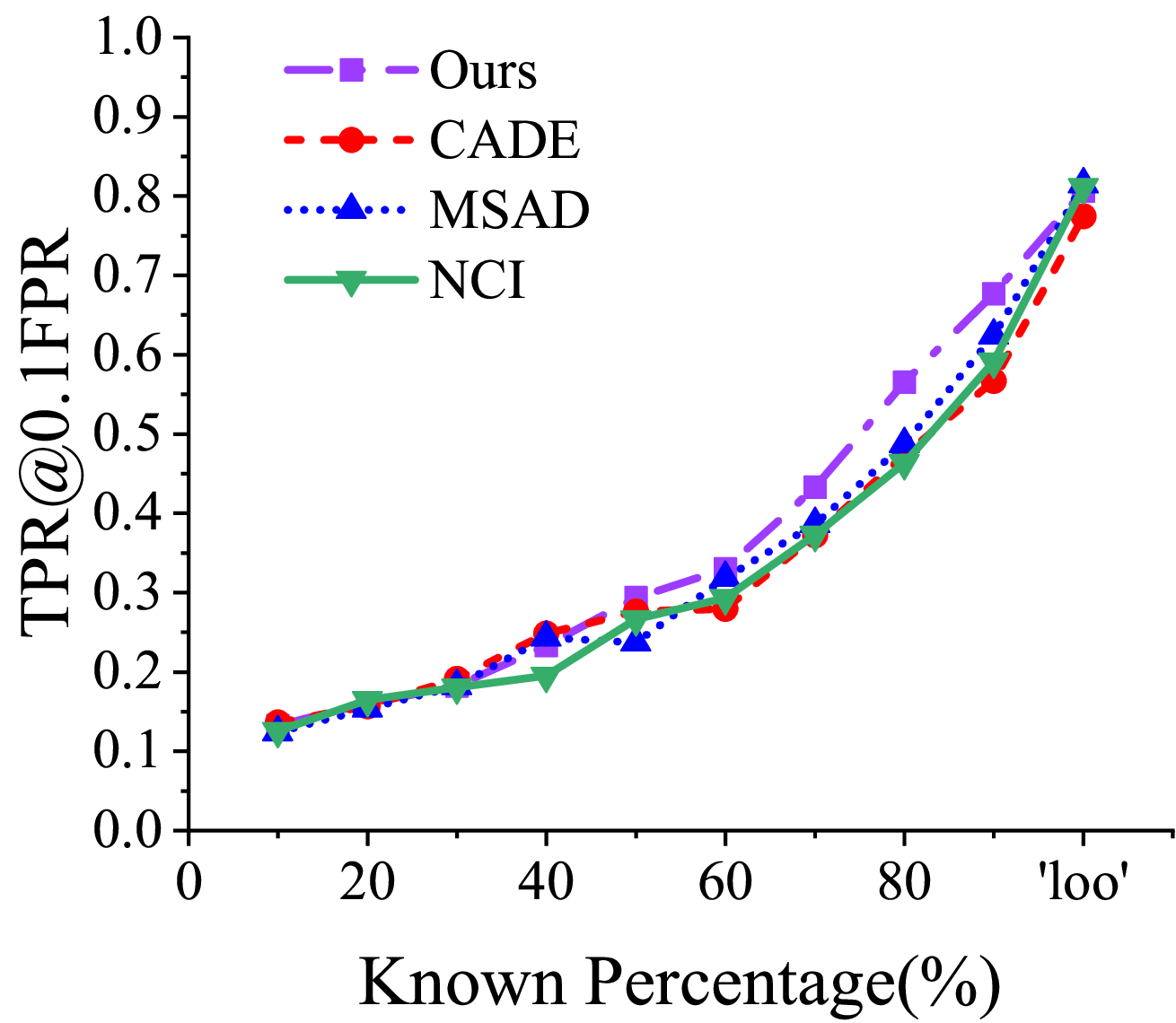}
    }
    \subfloat[STL-10]{
    \includegraphics[width=0.22\linewidth]{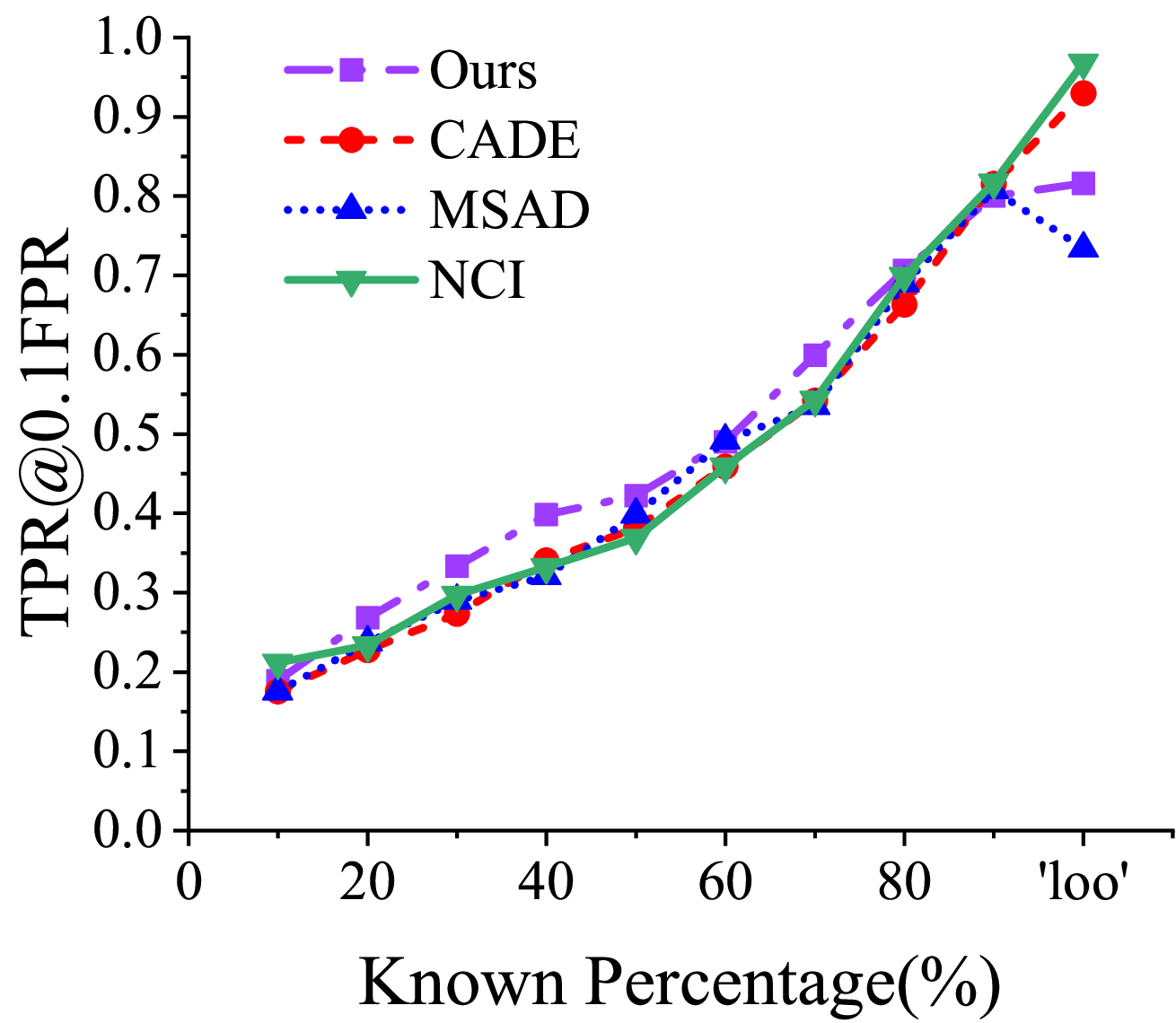}
    }
    \subfloat[Fashion-MNIST]{
    \includegraphics[width=0.22\linewidth]{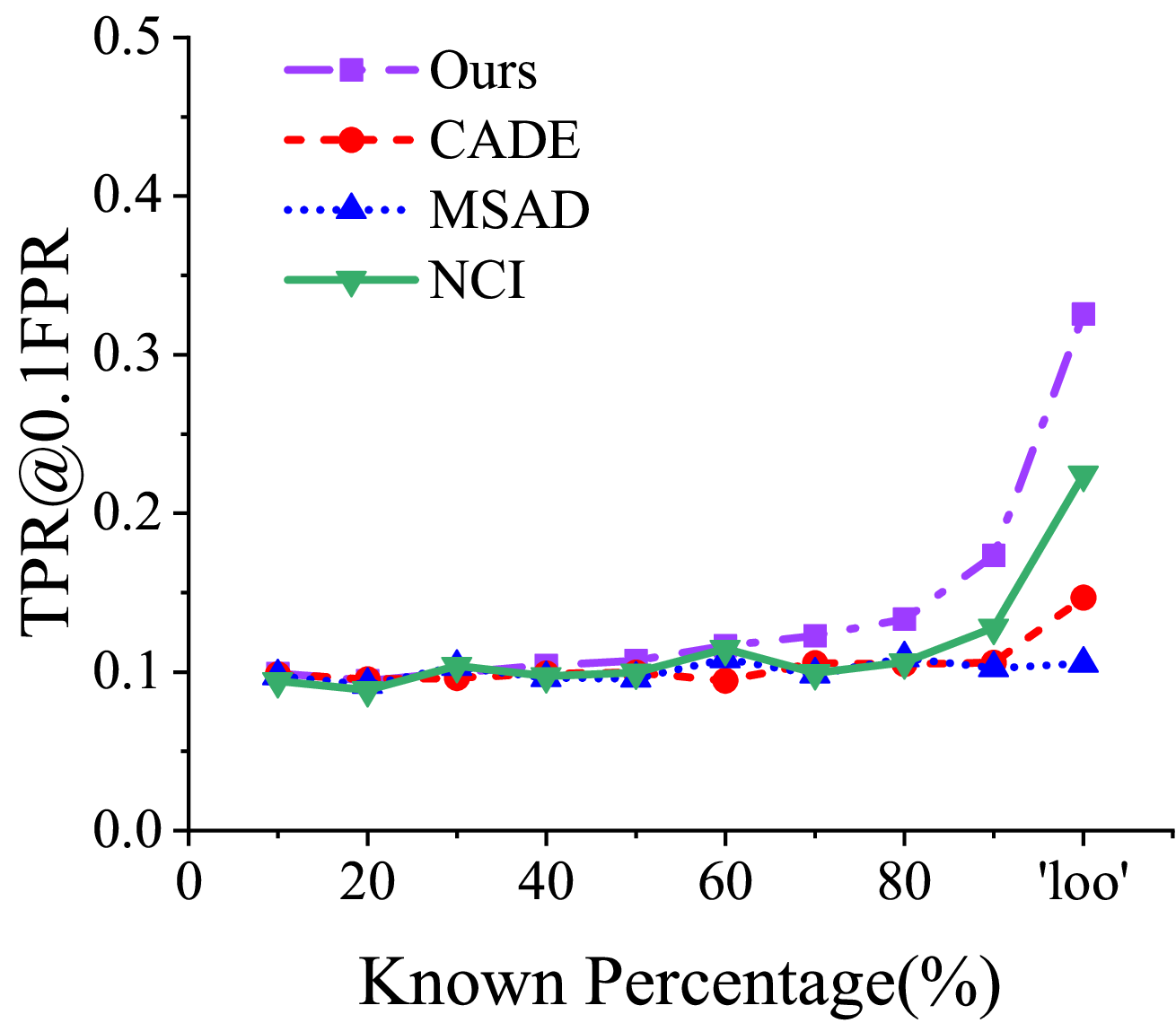}
    }
    \subfloat[Epsilon]{
    \includegraphics[width=0.22\linewidth]{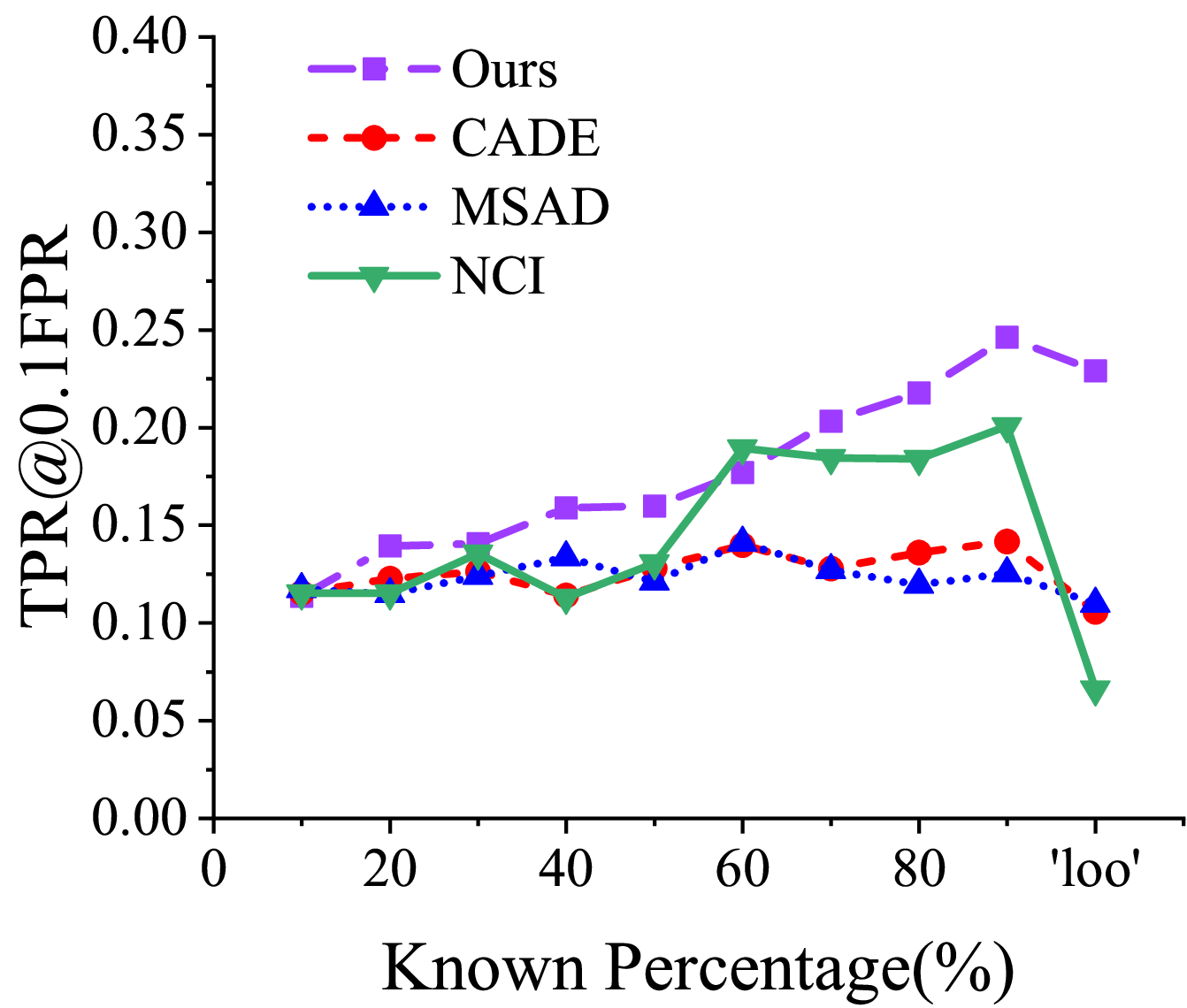}
    }
    \caption{Black-box attack performance (TPR@0.1FPR) under varying known feature proportions. The x-axis represents the percentage of known features, and the y-axis shows the corresponding attack TPR when TPR=0.1. Each curve corresponds to one of the four integrated anomaly detection methods within our attack framework. We denote the case where only one feature is unknown as “loo”.}
    \label{fig:b_tpr}
\end{figure*}

Here we preset the performance of our black-box attack. To obtain the estimated gradient, we sample 100 random perturbation with $\epsilon=0.01$ for each data. All other settings remain identical to those used in the white-box attack. The results are shown in Figure \ref{fig:b_auc} and Figure \ref{fig:b_tpr}.

Overall, the black-box attack-implemented via zeroth-order gradient estimation closely matches the white-box performance on four datasets in terms of both metrics. As the number of known features increases, the attack performance consistently improves. When 90\% of the features are known, the black-box attack achieves slightly lower performance than the white-box attack on CIFAR-10 and STL-10. This gap is more noticeable when considering TPR@0.1FPR. On Fashion-MNIST, although a more noticeable performance drop is observed, the attack remains within a practically usable range. In contrast, the difference between black-box and white-box settings on the Epsilon dataset is minimal. Additionally, it can be observed that, in the second stage of the black-box attack, using our proposed method yields a higher TPR at low FPR.

Notably, the performance degradation in the leave-one-out setting is greatly mitigated under the black-box formulation. A likely explanation is that the reconstruction problem reduces to a very low-dimensional search (one unknown feature), where finite-difference estimates act as a smoothing operator and produce more stable update directions than the raw white-box gradient. These findings indicate that zeroth-order estimation provides a practical and, in some cases, advantageous alternative to white-box gradients.

In summary, although the black-box setting results in a modest reduction in attack performance, the attack remains effective in extracting privacy information from the target model. Moreover, when considered together with partial feature availability, the black-box scenario further broadens the practical scope of membership inference attack, making it more realistic and applicable in real-world settings.

\subsection{Case Study}\label{sec:case_study}

We now present a realistic scenario to demonstrate how our proposed attack can be applied in a real-world context. Consider a small community where a local hospital provides medical services to most residents. The hospital has developed a patient risk prediction system based on historical records of diabetic patients to forecast the likelihood of readmission.

Suppose there is an adversary within a community who aims to infer private information about a specific individual named Alice. The attacker is assumed to know certain publicly accessible demographic information about Alice, such as her race, age, and gender, which helps narrow down the population scope. Using this partial information and the PFMI attack, the attacker successfully determines that Alice was part of the training dataset of the hospital. Since all training records in the dataset correspond to diabetic patients, the attacker can reasonably conclude that Alice is very likely a diabetic patient, thereby exposing sensitive medical information.

To simulate this scenario, we use the UCI-Diabetes dataset. We select 10 features for model training and designate race, age, and gender, common demographic attributes easily accessible to an attacker, as known features. Other experimental settings remain consistent with previous sections. The attack results are shown in Table \ref{tab:case_study}, demonstrating that our method can effectively compromise individual privacy in practical settings.

\begin{table}[!t]
\caption{AUC performance of case study.}
\label{tab:case_study}
\centering
\begin{tabular}{ccccc}
    \hline
    &Ours&CADE&MSAD&NCI  \\
    \hline
     AUC& 0.78&0.66&0.75&0.75\\
    \hline
\end{tabular}
\end{table}

\section{Discussion}\label{sec:discussion}


\subsection{Relationship to Other Attacks}
\textbf{Relationship to Property Existence Attack}: Property existence attack\cite{chaudhari2023snap}, as a special case of distribution inference attack\cite{chaudhari2023snap,suri2023dissecting}, aims to determine whether a given feature combination exists in the training set. The only difference from our attack lies in whether the attacker has access to the unknown features. Clearly, an attacker capable of performing PFMI can easily carry out a property existence attack, but it is difficult for a property existence attacker to perform PFMI because of the absence of some features.

\textbf{Relationship to Model Inversion Attack}: A common premise of existing model inversion attacks is that the adversary knows the target to be present in the training set\cite{dibbo2023model,kahla2022label,an2022mirror,annamalai2024linear,mehnaz2022your,kabir2025disparate}. Even though the membership status of the target is unknown in the attack scenario considered in this paper, the first stage of our attack can still be regarded as a form of model inversion attack. Leveraging existing model inversion techniques thus represents a feasible approach for improving performance.

\subsection{Limitation and Future Work}
While our proposed method demonstrates strong attack performance, there remain several directions worth exploring in future work.

First, the effectiveness of multi-round optimization could be further improved by adopting more advanced and efficient optimization strategies, such as incorporating momentum or adaptive step-size methods. These techniques may help stabilize the reconstruction process and enhance attack performance under limited query budgets.

Second, it is important to re-evaluate the effectiveness of existing defense mechanisms in the context studied in this work. For example, machine unlearning\cite{bourtoule2021machine,liu2025prototype} has been proposed as an effective defense against inference attacks by updating model parameters to forget a specific sample. However, even after forgetting, the model may still retain partial feature-level memorization, which could be exploited by attackers through PFMI to infer sensitive information. Other methods that have also been proven effective in defending against membership inference attacks, such as overfitting mitigation\cite{srivastava2014dropout} and differential privacy\cite{dwork2006differential,abadi2016deep}, also require further validation in the context of this work.

\section{Conclusion}\label{sec:conclusion}
In this paper, we study membership inference attack under the setting where some feature values are missing. To address this scenario, we introduce a practical two-stage attack framework named MRAD which is applicable to both white-box and black-box settings. Our framework is compatible with arbitrary anomaly detection methods, enabling attackers to flexibly conduct attacks under various knowledge settings.

We conduct extensive experiments on multiple datasets to evaluate the effectiveness of our method. The results show that our framework consistently delivers strong performance across a wide range of known feature percentages.  Remarkably, even when 60\% of the features are missing, the attack still achieves an AUC of 0.75 on STL-10. The attack tests in the black-box scenario also demonstrate that our attack can successfully extract training data privacy, even without access to the model parameters. In addition, we use SHAP to measure feature importance and experimentally confirm that the importance of the known features has a significant impact on the attack’s performance. Finally, through a case study, we simulate a real‑world application scenario to further highlight the practical relevance of our attack method.

Our findings demonstrate that membership inference attacks can be mounted using publicly available partial data, even when private features of a sample are protected. This highlights a new dimension in privacy risk assessment.

\bibliographystyle{IEEEtran}

\bibliography{reference}

@inproceedings{shokri2017membership,
  title={Membership inference attacks against machine learning models},
  author={Shokri, Reza and Stronati, Marco and Song, Congzheng and Shmatikov, Vitaly},
  booktitle={2017 IEEE symposium on security and privacy (SP)},
  pages={3--18},
  year={2017},
  organization={IEEE}
}

@inproceedings{hui2021practical,
  author = {Hui, Bo and Yang, Yuchen and Yuan, Haolin and Burlina, Philippe and Gong, Neil Zhenqiang and Cao, Yinzhi},
  title = {Practical Blind Membership Inference Attack via Differential Comparisons},
  booktitle = {Proceedings of the Network and Distributed System Security Symposium (NDSS'21)},
  year = {2021}, 
  month = {February},
}

@inproceedings{peng2024oslo,
  title={OSLO: one-shot label-only membership inference attacks},
  author={Peng, Yuefeng and Roh, Jaechul and Maji, Subhransu and Houmansadr, Amir},
  booktitle={Proceedings of the 38th International Conference on Neural Information Processing Systems},
  pages={62310--62333},
  year={2024}
}

@inproceedings{chaudhari2023snap,
  title={SNAP: Efficient extraction of private properties with poisoning},
  author={Chaudhari, Harsh and Abascal, John and Oprea, Alina and Jagielski, Matthew and Tramer, Florian and Ullman, Jonathan},
  booktitle={2023 IEEE Symposium on Security and Privacy (SP)},
  pages={400--417},
  year={2023},
  organization={IEEE}
}

@inproceedings{suri2023dissecting,
  title={Dissecting distribution inference},
  author={Suri, Anshuman and Lu, Yifu and Chen, Yanjin and Evans, David},
  booktitle={2023 IEEE Conference on Secure and Trustworthy Machine Learning (SaTML)},
  pages={150--164},
  year={2023},
  organization={IEEE}
}

@inproceedings{song2021systematic,
  title={Systematic evaluation of privacy risks of machine learning models},
  author={Song, Liwei and Mittal, Prateek},
  booktitle={30th USENIX Security Symposium (USENIX Security 21)},
  pages={2615--2632},
  year={2021}
}

@inproceedings{carlini2022membership,
  title={Membership inference attacks from first principles},
  author={Carlini, Nicholas and Chien, Steve and Nasr, Milad and Song, Shuang and Terzis, Andreas and Tramer, Florian},
  booktitle={2022 IEEE symposium on security and privacy (SP)},
  pages={1897--1914},
  year={2022},
  organization={IEEE}
}

@inproceedings{choquette2021label,
  title={Label-only membership inference attacks},
  author={Choquette-Choo, Christopher A and Tramer, Florian and Carlini, Nicholas and Papernot, Nicolas},
  booktitle={International conference on machine learning},
  pages={1964--1974},
  year={2021},
  organization={PMLR}
}

@inproceedings{li2021membership,
  title={Membership leakage in label-only exposures},
  author={Li, Zheng and Zhang, Yang},
  booktitle={Proceedings of the 2021 ACM SIGSAC Conference on Computer and Communications Security},
  pages={880--895},
  year={2021}
}

@inproceedings{wang2024property,
  title={Property existence inference against generative models},
  author={Wang, Lijin and Wang, Jingjing and Wan, Jie and Long, Lin and Yang, Ziqi and Qin, Zhan},
  booktitle={33rd USENIX Security Symposium (USENIX Security 24)},
  pages={2423--2440},
  year={2024}
}

@article{wares2019data,
  title={Data stream mining: methods and challenges for handling concept drift},
  author={Wares, Scott and Isaacs, John and Elyan, Eyad},
  journal={SN Applied Sciences},
  volume={1},
  pages={1--19},
  year={2019},
  publisher={Springer}
}

@inproceedings{wan2024online,
  title={Online Drift Detection with Maximum Concept Discrepancy},
  author={Wan, Ke and Liang, Yi and Yoon, Susik},
  booktitle={Proceedings of the 30th ACM SIGKDD Conference on Knowledge Discovery and Data Mining},
  pages={2924--2935},
  year={2024}
}

@inproceedings{yang2021cade,
  title={$\{$CADE$\}$: Detecting and explaining concept drift samples for security applications},
  author={Yang, Limin and Guo, Wenbo and Hao, Qingying and Ciptadi, Arridhana and Ahmadzadeh, Ali and Xing, Xinyu and Wang, Gang},
  booktitle={30th USENIX Security Symposium (USENIX Security 21)},
  pages={2327--2344},
  year={2021}
}

@inproceedings{reiss2023mean,
  title={Mean-shifted contrastive loss for anomaly detection},
  author={Reiss, Tal and Hoshen, Yedid},
  booktitle={Proceedings of the AAAI Conference on Artificial Intelligence},
  volume={37},
  number={2},
  pages={2155--2162},
  year={2023}
}

@inproceedings{liu2025detecting,
  title={Detecting out-of-distribution through the lens of neural collapse},
  author={Liu, Litian and Qin, Yao},
  booktitle={Proceedings of the Computer Vision and Pattern Recognition Conference},
  pages={15424--15433},
  year={2025}
}

@article{zhang2023openood,
  title={OpenOOD v1.5: Enhanced Benchmark for Out-of-Distribution Detection},
  author={Zhang, Jingyang and Yang, Jingkang and Wang, Pengyun and Wang, Haoqi and Lin, Yueqian and Zhang, Haoran and Sun, Yiyou and Du, Xuefeng and Li, Yixuan and Liu, Ziwei and Chen, Yiran and Li, Hai},
  journal={arXiv preprint arXiv:2306.09301},
  year={2023}
}

@inproceedings{park2023nearest,
  title={Nearest neighbor guidance for out-of-distribution detection},
  author={Park, Jaewoo and Jung, Yoon Gyo and Teoh, Andrew Beng Jin},
  booktitle={Proceedings of the IEEE/CVF international conference on computer vision},
  pages={1686--1695},
  year={2023}
}

@inproceedings{salem2023sok,
  title={SoK: Let the privacy games begin! A unified treatment of data inference privacy in machine learning},
  author={Salem, Ahmed and Cherubin, Giovanni and Evans, David and K{\"o}pf, Boris and Paverd, Andrew and Suri, Anshuman and Tople, Shruti and Zanella-B{\'e}guelin, Santiago},
  booktitle={2023 IEEE Symposium on Security and Privacy (SP)},
  pages={327--345},
  year={2023},
  organization={IEEE}
}

@inproceedings{sablayrolles2019white,
  title={White-box vs black-box: Bayes optimal strategies for membership inference},
  author={Sablayrolles, Alexandre and Douze, Matthijs and Schmid, Cordelia and Ollivier, Yann and J{\'e}gou, Herv{\'e}},
  booktitle={International Conference on Machine Learning},
  pages={5558--5567},
  year={2019},
  organization={PMLR}
}

@inproceedings{fredrikson2015model,
  title={Model inversion attacks that exploit confidence information and basic countermeasures},
  author={Fredrikson, Matt and Jha, Somesh and Ristenpart, Thomas},
  booktitle={Proceedings of the 22nd ACM SIGSAC conference on computer and communications security},
  pages={1322--1333},
  year={2015}
}

@inproceedings{annamalai2024linear,
  title={A linear reconstruction approach for attribute inference attacks against synthetic data},
  author={Annamalai, Meenatchi Sundaram Muthu Selva and Gadotti, Andrea and Rocher, Luc},
  booktitle={33rd USENIX Security Symposium (USENIX Security 24)},
  pages={2351--2368},
  year={2024}
}

@inproceedings{mehnaz2022your,
  title={Are your sensitive attributes private? novel model inversion attribute inference attacks on classification models},
  author={Mehnaz, Shagufta and Dibbo, Sayanton V and Kabir, Ehsanul and Li, Ninghui and Bertino, Elisa},
  booktitle={31st USENIX Security Symposium (USENIX Security 22)},
  pages={4579--4596},
  year={2022}
}

@inproceedings{yeom2018privacy,
  title={Privacy risk in machine learning: Analyzing the connection to overfitting},
  author={Yeom, Samuel and Giacomelli, Irene and Fredrikson, Matt and Jha, Somesh},
  booktitle={2018 IEEE 31st computer security foundations symposium (CSF)},
  pages={268--282},
  year={2018},
  organization={IEEE}
}

@incollection{NIPS2017_7062,
title = {A Unified Approach to Interpreting Model Predictions},
author = {Lundberg, Scott M and Lee, Su-In},
booktitle = {Advances in Neural Information Processing Systems 30},
editor = {I. Guyon and U. V. Luxburg and S. Bengio and H. Wallach and R. Fergus and S. Vishwanathan and R. Garnett},
pages = {4765--4774},
year = {2017},
publisher = {Curran Associates, Inc.},
url = {http://papers.nips.cc/paper/7062-a-unified-approach-to-interpreting-model-predictions.pdf}
}

@inproceedings{abadi2016deep,
  title={Deep learning with differential privacy},
  author={Abadi, Martin and Chu, Andy and Goodfellow, Ian and McMahan, H Brendan and Mironov, Ilya and Talwar, Kunal and Zhang, Li},
  booktitle={Proceedings of the 2016 ACM SIGSAC conference on computer and communications security},
  pages={308--318},
  year={2016}
}

@article{srivastava2014dropout,
  title={Dropout: a simple way to prevent neural networks from overfitting},
  author={Srivastava, Nitish and Hinton, Geoffrey and Krizhevsky, Alex and Sutskever, Ilya and Salakhutdinov, Ruslan},
  journal={The journal of machine learning research},
  volume={15},
  number={1},
  pages={1929--1958},
  year={2014},
  publisher={JMLR. org}
}

@inproceedings{hui2021trajnet,
  title={Trajnet: A trajectory-based deep learning model for traffic prediction},
  author={Hui, Bo and Yan, Da and Chen, Haiquan and Ku, Wei-Shinn},
  booktitle={Proceedings of the 27th ACM SIGKDD Conference on Knowledge Discovery \& Data Mining},
  pages={716--724},
  year={2021}
}

@article{chen2024end,
  title={End-to-end autonomous driving: Challenges and frontiers},
  author={Chen, Li and Wu, Penghao and Chitta, Kashyap and Jaeger, Bernhard and Geiger, Andreas and Li, Hongyang},
  journal={IEEE Transactions on Pattern Analysis and Machine Intelligence},
  year={2024},
  publisher={IEEE}
}

@article{bilionis2025disparate,
  title={Disparate Model Performance and Stability in Machine Learning Clinical Support for Diabetes and Heart Diseases},
  author={Bilionis, Ioannis and Berrios, Ricardo C and Fernandez-Luque, Luis and Castillo, Carlos},
  journal={AMIA Summits on Translational Science Proceedings},
  volume={2025},
  pages={95},
  year={2025}
}

@inproceedings{ye2022enhanced,
  title={Enhanced membership inference attacks against machine learning models},
  author={Ye, Jiayuan and Maddi, Aadyaa and Murakonda, Sasi Kumar and Bindschaedler, Vincent and Shokri, Reza},
  booktitle={Proceedings of the 2022 ACM SIGSAC Conference on Computer and Communications Security},
  pages={3093--3106},
  year={2022}
}

@inproceedings{zarifzadeh2024low,
  title={Low-Cost High-Power Membership Inference Attacks},
  author={Zarifzadeh, Sajjad and Liu, Philippe and Shokri, Reza},
  booktitle={International Conference on Machine Learning},
  pages={58244--58282},
  year={2024},
  organization={PMLR}
}

@article{krizhevsky2009learning,
  title={Learning multiple layers of features from tiny images},
  author={Krizhevsky, Alex and Hinton, Geoffrey and others},
  year={2009},
  publisher={Toronto, ON, Canada}
}

@article{xiao2017fashion,
  title={Fashion-mnist: a novel image dataset for benchmarking machine learning algorithms},
  author={Xiao, Han and Rasul, Kashif and Vollgraf, Roland},
  journal={arXiv preprint arXiv:1708.07747},
  year={2017}
}

@inproceedings{coates2011analysis,
  title={An analysis of single-layer networks in unsupervised feature learning},
  author={Coates, Adam and Ng, Andrew and Lee, Honglak},
  booktitle={Proceedings of the fourteenth international conference on artificial intelligence and statistics},
  pages={215--223},
  year={2011},
  organization={JMLR Workshop and Conference Proceedings}
}

@misc{pascal_large_scale,
  title        = {Pascal Large Scale Learning Challenge},
  author       = {{PASCAL}},
  year         = {2008},
  howpublished = {\url{https://www.csie.ntu.edu.tw/\~{}cjlin/libsvmtools/datasets/binary.html}},
}

@inproceedings{chi2024shadow,
  title={Shadow-free membership inference attacks: recommender systems are more vulnerable than you thought},
  author={Chi, Xiaoxiao and Zhang, Xuyun and Wang, Yan and Qi, Lianyong and Beheshti, Amin and Xu, Xiaolong and Choo, Kim-Kwang Raymond and Wang, Shuo and Hu, Hongsheng},
  booktitle={Proceedings of the Thirty-Third International Joint Conference on Artificial Intelligence},
  pages={5781--5789},
  year={2024}
}

@inproceedings{wang2023link,
  title={Link membership inference attacks against unsupervised graph representation learning},
  author={Wang, Xiuling and Wang, Wendy Hui},
  booktitle={Proceedings of the 39th Annual Computer Security Applications Conference},
  pages={477--491},
  year={2023}
}

@inproceedings{tao2025range,
  title={Range membership inference attacks},
  author={Tao, Jiashu and Shokri, Reza},
  booktitle={2025 IEEE Conference on Secure and Trustworthy Machine Learning (SaTML)},
  pages={346--361},
  year={2025},
  organization={IEEE}
}

@article{hu2022membership,
  title={Membership inference attacks on machine learning: A survey},
  author={Hu, Hongsheng and Salcic, Zoran and Sun, Lichao and Dobbie, Gillian and Yu, Philip S and Zhang, Xuyun},
  journal={ACM Computing Surveys (CSUR)},
  volume={54},
  number={11s},
  pages={1--37},
  year={2022},
  publisher={ACM New York, NY}
}

@inproceedings{dibbo2023model,
  title={Model inversion attack with least information and an in-depth analysis of its disparate vulnerability},
  author={Dibbo, Sayanton V and Chung, Dae Lim and Mehnaz, Shagufta},
  booktitle={2023 IEEE Conference on Secure and Trustworthy Machine Learning (SaTML)},
  pages={119--135},
  year={2023},
  organization={IEEE}
}

@inproceedings{kahla2022label,
  title={Label-only model inversion attacks via boundary repulsion},
  author={Kahla, Mostafa and Chen, Si and Just, Hoang Anh and Jia, Ruoxi},
  booktitle={Proceedings of the IEEE/CVF conference on computer vision and pattern recognition},
  pages={15045--15053},
  year={2022}
}

@inproceedings{an2022mirror,
  title={Mirror: Model inversion for deep learning network with high fidelity},
  author={An, Shengwei and Tao, Guanhong and Xu, Qiuling and Liu, Yingqi and Shen, Guangyu and Yao, Yuan and Xu, Jingwei and Zhang, Xiangyu},
  booktitle={Proceedings of the 29th Network and Distributed System Security Symposium},
  year={2022}
}

@inproceedings{dwork2006differential,
  title={Differential privacy},
  author={Dwork, Cynthia},
  booktitle={International colloquium on automata, languages, and programming},
  pages={1--12},
  year={2006},
  organization={Springer}
}

@inproceedings{kabir2025disparate,
  title={Disparate Privacy Vulnerability: Targeted Attribute Inference Attacks and Defenses},
  author={Kabir, Ehsanul and Craig, Lucas and Mehnaz, Shagufta},
  booktitle={34st USENIX Security Symposium (USENIX Security 25)},
  year={2025}
}

@article{wibisono2012finite,
  title={Finite sample convergence rates of zero-order stochastic optimization methods},
  author={Wibisono, Andre and Wainwright, Martin J and Jordan, Michael and Duchi, John C},
  journal={Advances in Neural Information Processing Systems},
  volume={25},
  year={2012}
}

@inproceedings{bourtoule2021machine,
  title={Machine unlearning},
  author={Bourtoule, Lucas and Chandrasekaran, Varun and Choquette-Choo, Christopher A and Jia, Hengrui and Travers, Adelin and Zhang, Baiwu and Lie, David and Papernot, Nicolas},
  booktitle={2021 IEEE symposium on security and privacy (SP)},
  pages={141--159},
  year={2021},
  organization={IEEE}
}

@inproceedings{liu2025prototype,
  title={Prototype Surgery: Tailoring Neural Prototypes via Soft Labels for Efficient Machine Unlearning},
  author={Liu, Gaoyang and Wang, Xijie and Wang, Zixiong and Wang, Chen and Abdelmoniem, Ahmed M and Wang, Desheng},
  booktitle={Proceedings of the 2025 ACM SIGSAC Conference on Computer and Communications Security},
  pages={2804--2817},
  year={2025}
}

@article{liu2023membership,
  title={Membership Inference Attacks Against Machine Learning Models via Prediction Sensitivity},
  author={Liu, Lan and Wang, Yi and Liu, Gaoyang and Peng, Kai and Wang, Chen},
  journal={IEEE Transactions on Dependable and Secure Computing},
  volume={20},
  number={3},
  pages={2341--2347},
  year={2023},
  publisher={IEEE}
}

@article{fu2025unlocking,
  title={Unlocking Generative Priors: A New Membership Inference Framework for Diffusion Models},
  author={Fu, Xiaomeng and Wang, Xi and Li, Qiao and Liu, Jin and Dai, Jiao and Han, Jizhong and Gao, Xingyu},
  journal={IEEE Transactions on Information Forensics and Security},
  year={2025},
  publisher={IEEE}
}

@article{galichin2025glira,
  title={Glira: Black-box membership inference attack via knowledge distillation},
  author={Galichin, Andrey V and Pautov, Mikhail and Zhavoronkin, Alexey and Rogov, Oleg Y and Oseledets, Ivan},
  journal={IEEE Transactions on Information Forensics and Security},
  year={2025},
  publisher={IEEE}
}

@article{hu2025unveiling,
  title={Unveiling Privacy Risks in the Long Tail: Membership Inference in Class Skewness},
  author={Hu, Hailong and Pang, Jun and Li, Yantao and Qin, Huafeng},
  journal={IEEE Transactions on Information Forensics and Security},
  year={2025},
  publisher={IEEE}
}

@article{liu2024gradient,
  title={Gradient-Leaks: Enabling Black-Box Membership Inference Attacks Against Machine Learning Models},
  author={Liu, Gaoyang and Xu, Tianlong and Zhang, Rui and Wang, Zixiong and Wang, Chen and Liu, Ling},
  journal={IEEE Transactions on Information Forensics and Security},
  volume={19},
  pages={427--440},
  year={2024},
  publisher={IEEE}
}

\end{document}